\documentclass[final]{cvpr}

\usepackage{times}
\usepackage{epsfig}
\usepackage{graphicx}
\usepackage{amsmath}
\usepackage{amssymb}
\usepackage{enumitem}
\usepackage{subcaption}
\usepackage{mathtools, nccmath}
\usepackage{amssymb,xcolor}
\usepackage{multicol}
\usepackage{multirow}
\usepackage{array,microtype}
\usepackage{tabularx}
\usepackage{graphbox}
\usepackage{array}
\usepackage{makecell}
\usepackage{placeins}
\usepackage{stfloats}

\usepackage[pagebackref=true,breaklinks=true,colorlinks,bookmarks=false]{hyperref}
\renewcommand{\vec}[1]{\mathbf{#1}}
\newcolumntype{P}[1]{>{\centering\arraybackslash}p{#1}}
\newcolumntype{C}{>{\centering\arraybackslash}X}

\newlength{\subcolumnwidth}

\newcommand{\nextsubcolumn}[1][]{%
  \cr\noalign{\hfill}
  \if\relax\detokenize{#1}\relax\else\hsize=#1\setlength{\subcolumnwidth}{\hsize}\fi
}

\newcommand{\mnamefull}[0]{{\bf SCG Object Localiser}}
\newcommand{\mname}[0]{{\bf SCG-OL}}

\def\confYear{CVPR 2022}

\begin{document}

\title{Spatial Commonsense Graph for Object Localisation in Partial Scenes}
\author{
Francesco Giuliari\textsuperscript{1,2}
\quad
Geri Skenderi\textsuperscript{3}
\quad
Marco Cristani\textsuperscript{3}
\quad
Yiming Wang\textsuperscript{1,4}
\quad
Alessio Del Bue\textsuperscript{1}
\vspace{1ex}
\\
\textsuperscript{1}Istituto Italiano di Tecnologia (IIT) \quad
\textsuperscript{2}University of Genova \quad 
\textsuperscript{3}University of Verona \quad \\
\textsuperscript{4}Fondazione Bruno Kessler (FBK)
}

\maketitle

\begin{abstract}

We solve object localisation in partial scenes, a new problem of estimating the unknown position of an object (e.g. where is the bag?) given a partial 3D scan of a scene. The proposed solution is based on a novel scene graph model, the Spatial Commonsense Graph (SCG), where objects are the nodes and edges define pairwise distances between them, enriched by concept nodes and relationships from a commonsense knowledge base. This allows SCG to better generalise its spatial inference over unknown 3D scenes. The SCG is used to estimate the unknown position of the target object in two steps:
first, we feed the SCG into a novel Proximity Prediction Network, a graph neural network that uses attention to perform distance prediction between the node representing the target object and the nodes representing the observed objects in the SCG; second, we propose a Localisation Module based on circular intersection to estimate the object position using all the predicted pairwise distances in order to be independent of any reference system. We create a new dataset of partially reconstructed scenes to benchmark our method and baselines for object localisation in partial scenes, where our proposed method achieves the best localisation performance.
\footnote{Code and Dataset are available here: \url{https://github.com/FGiuliari/SpatialCommonsenseGraph-Dataset}} %

\end{abstract}
\section{Introduction}
\label{sec:intro}

\begin{figure}[t!]
	\centering
	\includegraphics[width=0.8\linewidth]{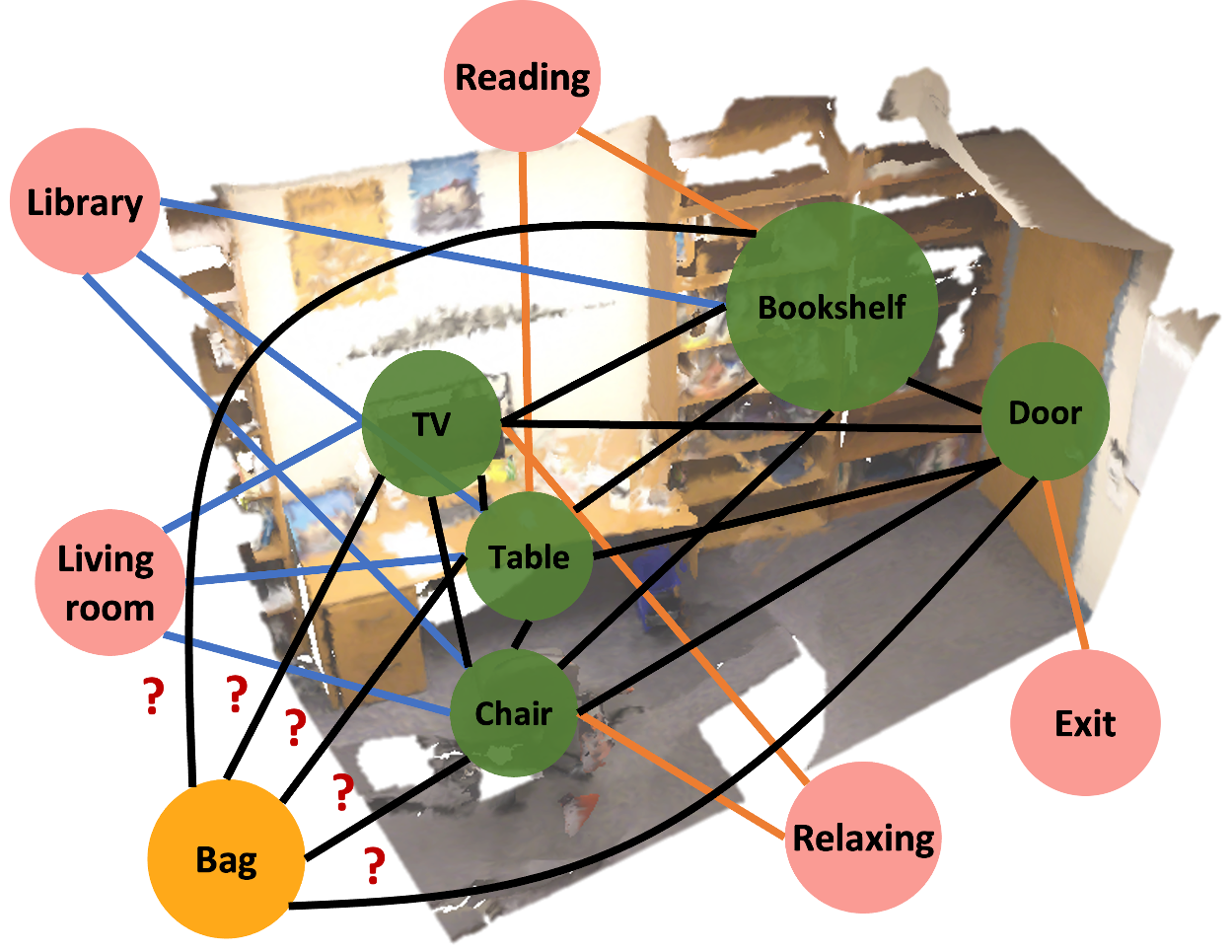}
	\vspace{-0.3cm}
	\caption{Given a set of objects (indicated in the {\color{green} \textbf{green}} circles) in a partially known scene, we aim at estimating the position of a target object (indicated in the {\color{orange} \textbf{orange}} circle). We treat this localisation problem as an edge prediction problem by constructing a novel scene graph representation, the Spatial Commonsense Graph (SCG), that contains both the spatial knowledge extracted from the reconstructed scene, i.e. the proximity (\textbf{black} edges) and the commonsense knowledge represented by a set of relevant concepts (indicated in the {\color{pink} \textbf{pink}} circles) connected by relationships, e.g. {\it UsedFor} ({\color{orange} \textbf{orange}} edges) and {\it AtLocation} ({\color{blue} \textbf{blue}} edges).}
	\label{fig:teaser}
	\vspace{-0.4cm}
\end{figure}

The localisation of unobserved objects given a partial observation of a scene, as shown in Fig.~\ref{fig:teaser}, is a fundamental task that humans solve often in their everyday life. Such a task is useful for many automation applications, including domotics for assisting visually impaired humans to find everyday items~\cite{elmannai2017sensor}, visual search for embodied agents~\cite{batra2020objectnav}, and layout proposal for interior design~\cite{luo2020end}. Yet, object localisation in partial scenes has never been formally studied in the literature. 
We formalise the problem as the inference of the position of an arbitrary object in an unknown area of a scene based only on a partial observation of the scene.

Humans perform such an object localisation task by not only using the partially observed environment, but also by relying on the {\it commonsense} knowledge that is acquired during our lifetime experience. For example, by knowing that pillows are often close to beds (the {\it spatial} relationship), and that chairs and beds are often used for resting (the {\it affordance} relationship), one could infer the whereabouts of pillows even if only a bed and a chair were observed. In this paper, we question whether it is possible to computationally solve this task by injecting the commonsense knowledge within a scene graph representation \cite{krishnavisualgenome,gay2018visual,Wald20203Dssg}, so that a machine can also reasonably localise an object in the unseen part of scene, without the use of any visual/depth information. 

In this work, we propose a new scene graph representation, the \textbf{Spatial Commonsense Graph (SCG)}, having heterogeneous nodes and edges that embed the commonsense knowledge together with the spatial proximity of objects as measured in the partial 3D scan of the scene. The underlying intuition is that commonsense knowledge is extracted from an external knowledge base is not specific to any observed visual scene, and thus allows for a better generalisation, but at the cost of a coarser localisation. At the same time, the objects' arrangement in the known portion of the scene is useful in providing better pairwise object distances,  strengthening the estimate of the target object position. The main challenge here is devising a model that promotes the generalisation of commonsense while increasing the accuracy of the scene-specific metrics.

\begin{figure*}[t!]
	\centering
	\includegraphics[width=1\linewidth]{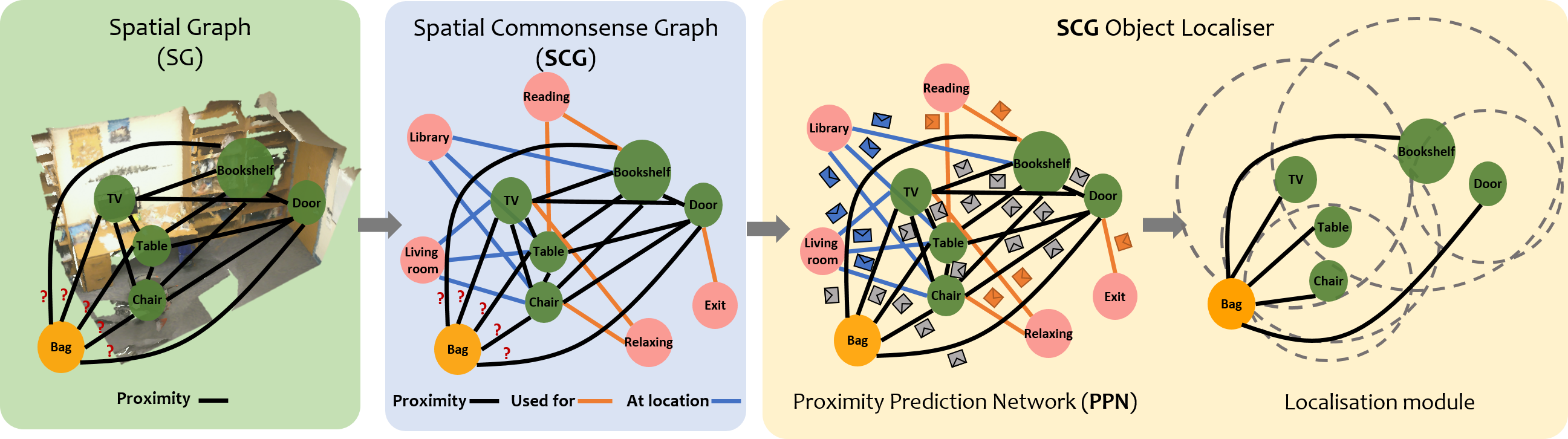}
	\vspace{-0.5cm}
	\caption{Overall architecture of our proposed approach. First, we construct a spatial commonsense graph (SCG) from the known scene by enriching the scene graph with concept relationships, resulting in edges of three types: {\it UsedFor} ({\color{orange} \textbf{orange}} edges), {\it AtLocation} ({\color{blue} \textbf{blue}} edges) and Proximity (\textbf{black} edges). The SCG is then fed into the Proximity Prediction Network (PPN) that performs message passing with attention to update the node features taking into consideration the heterogeneous edges. PPN then concatenates the node features of the target node and one of the scene object nodes and passes it through an MLP to predict the pairwise distance. The localisation module then uses the predicted pairwise distances to estimate the position of the target object within the area where most distances overlap.}
	
	\label{fig:architecture}
	\vspace{-0.1cm}
\end{figure*}

The proposed scene graph, as shown in Fig.~\ref{fig:architecture}, is first defined by nodes representing the known objects in the scene that are fully connected through edges representing the {\it proximity}, \emph{i.e.} the relative distance between a pair of objects. We call this spatial representation the Spatial Graph (SG) of the known partial 3D scan. Then, the SG is further expanded into the SCG by adding and connecting nodes that represent concepts through relevant commonsense relationships extracted from ConceptNet \cite{speer2018conceptnet}.

The SCG is instrumental to address the localisation problem. In this work, we propose a two-stage solution, dubbed \mnamefull~(\mname). First we predict the pairwise proximity between the target object node, having an unknown position, and each of the known object nodes through our graph-based \emph{Proximity Prediction Network} (PPN), formulating the task as an edge regression problem.  We then use our \emph{Localisation Module} to compute the position of the target based on the pairwise distances. The localisation module estimates the most probable position as the intersection of the circular areas defined by all pairwise object distances.  Note that by only using relative distances between pairs of objects, our model does not depend on the scene's reference frame, thus being considered agnostic to the coordinate system.

We also introduce a new dataset built from partial reconstructions of real-world indoor scenes using RGB-D sequences from ScanNet~\cite{dai2017scannet}, which we will use as a benchmark for this novel problem.
We construct the dataset to reflect different completeness levels of the reconstructed scenes. We define the evaluation protocol via a set of performance measures to quantify the localisation success and accuracy.

To summarise, our core contributions are the following:
\begin{itemize}[noitemsep,nolistsep]
    \item We identify a novel task of object localisation in partial scenes and propose a graph-based solution. We make available a new dataset and evaluation protocol, and show that our method achieves the best performance w.r.t. other comparing methods. %

    \item We propose a new heterogeneous scene graph, the \textbf{Spatial Commonsense Graph}, for an effective integration between the commonsense knowledge and the spatial scene, using attention-based message passing for the graph updates to prioritise the assimilation of knowledge relevant to the task. %
    
    \item We propose \mnamefull, a two-staged localisation solution that is agnostic to scene coordinates. The distances between the unseen object and all known objects are first estimated and then used for the localisation based on circular intersections.

\end{itemize}

\section{Related work}
\label{sec:sota}

We will cover prior work related to the inference of scene graphs, the current dataset used for experimental validation and the use of commonsense for spatial reasoning. %

\noindent\textbf{Scene graph modelling and inference.}
Scene graphs were initially used to describe images of scenes based on the elements they contained and how they were connected. The work of ~\cite{Johnson2015RetrievalGraph} showed that for certain applications, e.g. Image Retrieval, the abstraction of higher-level image concepts was improving the results compared to using the standard pixel space. Since then, scene graphs have been successfully used in many other tasks such as image captioning~\cite{xu2019scene, yang2019auto, Gu_2019_ICCV} and visual question answering~\cite{Shi_2019_CVPR, lee2019visual}.

Recently, the use of scene graphs has also been extended to the 3D domain, providing an efficient solution for 3D scene description. The 3D scene graph can vary from a simple representation of a scene and its content, in which the objects are nodes, and the spatial relationships between objects are the graph's  edges~\cite{gay2018visual, Wald20203Dssg,Wu_2021_Scenegraphfusion}; to a more complex hierarchical structure that describes the scene at different levels: from the image level with description about the scene from only a certain point of view, moving up to a higher level description of objects, rooms and finally buildings~\cite{armeni20193d}.
The work of \cite{Zhou2019SceneGraphNetNM} uses a scene graph to augment 3D indoor scenes with new objects matching their surroundings using a neural message passing approach.
A relatively similar task is indoor scene synthesis~\cite{wang19PlanIT}, in which the goal is to generate a new layout of a scene using a relation graph encoding objects as nodes and spatial/semantic relationships between objects as edges. A graph convolutional generative model  synthesises novel relation graphs and thus new layouts. In \cite{dhamo2021graphto3d} and \cite{luo2020end2end} the authors use a 3D scene graph to describe the object arrangement, they then modify the scene graph and try to generate a new scene. 

Like these works, we use an underlying scene representation, but unlike them we embed commonsense knowledge into the scene graph. In this way, our approach can better generalise to unseen rooms with unseen object arrangements by leveraging prior semantic knowledge.

\noindent \textbf{Dataset for Object Localisation.} Datasets existing in the literature are not suited for this type of object localisation task. For instance, Scene Synthesis datasets ~\cite{wang-synthesisprior} do not have enough variability in the scene structure, as all environments represented are of identical shape (rectangular) and of similar size. Moreover, the scenes mostly contains the same set of objects.
These characteristics lead to datasets that do not reflect the real world and cannot be used to train approaches to be deployed in real indoor environments. Another major limitation of existing datasets is their assumption that the entire layout of the room is known and all the objects lie within the boundaries of the observed part of the scene~\cite{wang19PlanIT, li2019grains}. This is typically not the case. In robotic applications like Visual Search, the robot only has partial information about the environment, that gets updated during navigation. In general, the searched object has to be found in the unexplored part of the scene, yet to be discovered. Our work is based on partially observed scenes and performs localisation without navigation.

\noindent\textbf{Embedding Commonsense Knowledge in Neural Networks.}
Commonsense reasoning focuses on imitating the high level reasoning employed by humans when solving tasks. Typically, we do not only use the information directly related to the task, but also rely on knowledge gained through prior experience. %
The field of Natural Language Processing,~\cite{faldu2021kibert} makes use of ConceptNet~\cite{speer2018conceptnet} to create richer, contextualised sentence embeddings with the BERT architecture~\cite{devlin2019bert}. In ~\cite{bao2014knowledge} , the authors utilise the knowledge graph Freebase (now Google Knowledge Graph) to enrich textual representations in a knowledge-based question answering system. In computer vision,~\cite{li2017incorporating} exploits commonsense knowledge using Dynamic Memory Networks for Visual Question Answering (VQA), stating it helps the network to reason beyond the image contents. In the scene graph generation task,~\cite{gu2019scene} exploits the ConceptNet ~\cite{speer2018conceptnet} knowledge graph to refine object and phrase features to improve the generalisation of the model. The authors state that the knowledge surrounding the subject of interest also benefits the inference of objects related to it, helping the model to generalise better and generate meaningful scene graphs. In this work, we exploit the commonsense knowledge to enrich a spatial scene representation used for predicting proximity among pairs of objects in a scene context.

\section{Spatial Commonsense Graph (SCG)}
\label{sec:graph_modelling}
Our model of the scene has the objective to embed commonsense knowledge into a geometric scene graph extracted from a partial scan of an area. 

As illustrated in Fig.~\ref{fig:architecture}, we construct the SCG with nodes that are  \textit{i)} object nodes including all the observed objects in the partially known environment and any target unseen object to be localised, or \textit{ii)} concept nodes that are retrieved from ConceptNet~\cite{speer2018conceptnet}. 
Each SCG is constructed on top of a Spatial Graph (SG) composed of object nodes that are fully connected. Each object node is further connected to concept nodes via the semantic relationships.
The edges of SCG are of three heterogeneous types:
\begin{itemize}[noitemsep,nolistsep]
    \item \emph{Proximity} relates the pairwise distances between \emph{all} the object nodes given the partial 3D scan; 
    \item \emph{AtLocation} is retrieved from ConceptNet, indicating which environment the objects are often located in; 
    \item \emph{UsedFor} is retrieved from ConceptNet, describing the common use of the objects.
\end{itemize}
The proximity edges connect all the objects nodes of the SCG in a fully connected manner, while the semantic \emph{AtLocation} and \emph{UsedFor} edges connect each object node with its related concept nodes that are queried from ConceptNet (e.g. \textit{bed AtLocation apartment} or \textit{bed UsedFor resting}). 
The two semantic edge types provide useful hints on how objects can be clustered in the physical space, thus benefiting the position inference of indoor objects.

We formulate SCG as an undirected graph that is composed by a set of nodes $\mathcal{H}=\{\vec h_i| \: i\in(0,N]\}$,
where $N = N_o+N_c$ is the total number of nodes in SCG with $N_o$ the number of the object nodes and $N_c$ the number of the concept nodes.
The $D$-vector $\vec h_i$ is the node's corresponding word embedding in NumberBatch~\cite{speer2017numberbatch} (i.e. $D = 300$). %
The edges are defined by the set  $\mathcal{E}=\{\vec e_{i,j}| \: {i,j\in(0,N], i\neq j \}}$, where $e_{i,j}$ is the edge between node $i$ and node $j$. Let $\mathcal{N}_i$ be the neighbouring nodes of node $i$ connected by any edge.
We use a 4-dimensional feature vector, i.e. $\vec e_{i,j} \in\mathbb{R}^{4}$, whose first three elements indicate the previously defined edge type in a one-hot manner while the last element is a scalar indicating the pairwise distance between two scene objects. 
Note that the distance is only measurable on the observed part of the 3D scan (i.e. between known object nodes). Otherwise, we initialise the distance value to $-1$ when the edges are \textit{AtLocation}, \textit{UsedFor}, or \textit{proximity} edges involving the unknown target object node. %

\section{SCG Object Localiser (SCG-OL)}
\label{sec:method}
We define a two-stage solution to address the task of localising the arbitrary unobserved target object using the  SCG. In the first stage, we propose a Proximity Prediction Network (PPN) on top of the SCG. PPN aims to predict the pairwise distances between the unseen target object and the objects in the partially known scene. In the second stage, our localisation module takes as input the set of pairwise distances and it outputs the position of the target object based on a probabilistic circular intersection.  The following sections provide more details regarding the Proximity Prediction Network and the Localisation module.
\subsection{Proximity Prediction Network}

The goal of the PPN is to predict all the pairwise distances between the unseen object and the observed scene objects. 
We utilise a variant of the Graph Transformer~\cite{shi-graphtransformer} and update the nodes iteratively over the heterogeneous edges, to allow effective fusion between the commonsense knowledge and the metric measurements.

The input to the network is the set of node features $\mathcal{H}$ and the output is a new set of node features $\mathcal{H}^{'}=\{\vec h_i'| \: i\in(0,N]\}$, with $\vec h_i'\in\mathbb{R}^D$. 
Each node $i$ in the graph is updated by aggregating the features of its neighbouring nodes $\mathcal{N}^i$ via two rounds
of message passing. The resulting $\vec h_i'$ forms a \emph{contextual} representation of its neighbourhood.

At each round of  message passing, we first learn the attention coefficient $\alpha_{i,j}$ using a graph based version of the scaled dot-product attention mechanism~\cite{shi-graphtransformer}, conditioned on each edge feature $\vec e_{i,j}$ from node $j$ to node $i$, and on both nodes' features, $ \vec h_i$ and $\vec h_j$ . This allows the network to understand how important each neighbour is for updating the node representation, which is performed in the following way:
\begin{align}
    \vec v_j & = {W_v}\vec h_j + b_v \label{eq:1} \\
    \hat{\vec h}_i & = {\sum_{j \in \mathcal{N}_i}}\alpha_{{ij}}(\vec v_{j} + e_{{i,j}}) \label{eq:2}
\end{align}
where $W_v, b_v$ represent respectively the weight matrix and bias used to calculate the value vector $\vec v$ for the scaled dot-product attention mechanism. %
The updated state $\vec h_i'$ is then defined as:
\begin{gather}
    \vec h_i' = ReLU(LNorm((1-\beta_i)\hat{\vec h}_i + \beta_i{W_r}\vec h_i + b_{r})) \label{eq:3}
\end{gather}
where $\beta_i$ is the output of a gated residual connection~\cite{shi-graphtransformer}, which prevents all the nodes from converging into indistinguishable features. The elements $W_r, b_r$ represent the weight matrix and bias respectively used in the linear transformation of $\vec h_i$.

After  message passing, we obtain the set of final node embeddings $\mathcal{H}^{*}=\{ \vec{h}_i^*| \: i\in(0,N]\}$, with $\vec h_i^*\in\mathbb{R}^{2D}= Concat(\vec h_i, \vec{h'}_i)$, where $Concat(\cdot)$ represents a concatenation operation.
This way, the final representation of each node contains both the original object embedding and the aggregated embedding of its context in the scene. 
Finally, we combine the features of the two nodes $\vec h_{i,t}^*=Concat(\vec{h}_i^*, \vec{h}_t^*)$ by concatenation, and predict the pairwise distances $\hat{d}_{i,t}$ between the target object node $t$ and the observed object node $i$ via fully connected layers.

\paragraph*{SCG-OL loss.} 

To train our PPN, we compute the Mean Square Error (MSE) between the predicted pairwise distances $\hat{d}_{i,t}$ of the object node $i$ and the target node $t$ and the set of ground-truth pairwise distances $d_{i,t}$ and use it as a loss:
\begin{equation}
    \mathcal{L}_{\text{MSE}}(\hat{d},d)=\frac{1}{N_o - 1}\sum_{i=1}^{N_o-1}{(\hat{d}_{i,t}-d_{i,t})}^2
    \label{loss:mse}
\end{equation}
Note that the class of the target object can have multiple instances in the unknown part of the scene, i.e. multiple ground-truth positions. Our method, as a localiser, uses the GT position of the instance that is closest to the predicted position for the computation of the MSE loss.

\subsection{Distances to position: Localisation module}

In the localisation module, we solve the problem of converting the set of predicted object-to-object distances to a single position $\hat{\vec p}_t$ in the space that defines the position of the searched object in a bird's eye view.
The distances $\hat{d}_{i,t}$ predicted by the PPN, and the known objects positions $\vec p_i$, can be used to define a set of circles of radius $\hat{d}_{i,t}$, centred in the positions $\vec p_i$. 
With perfect predictions, $\hat{\vec p}_t$ would be obtained as the point of intersection of all the circles. In this case we would need at least three known object nodes to unambiguously define $\hat{\vec p}_t$. For this reason, in this study we only consider instances with more than three or more known objects.

Let us define $\hat{\vec p}_t$ as the point in the space that minimises the squared distance from all the circles:
\begin{equation}
    \hat{\vec p_t}=\underset{\vec p_t}{\arg\,\min} \sum_{i}^{N_o-1}{ (\Vert \vec p_t - \vec p_i \Vert_2 - \hat{d}_i)^2}.
    \label{eq:localisation}
\end{equation}
While it is possible to obtain a closed form solution of Eq.~\ref{eq:localisation} via Linear Least Squares~\cite{Wang2015}, it would not be robust to noise in the measured distances, noise which is likely present in the PPN predictions. An alternative is to minimise this problem by brute force: we first subdivide the space into a grid and compute the sum of the residuals at each position. We then take the position with the lowest value and use it as an initial guess for the Nelder-Mead's simplex algorithm~\cite{neldermeanSimplex} to obtain the final estimate.

\section{Experiments}
\label{sec:exp}

We evaluate our proposed method on a new dataset of partially reconstructed indoor scenes. First, we provide the implementation details of our method followed by the evaluation metrics used for evaluation.

\noindent\textbf{Implementation Details.} We train our networks using the Adafactor optimiser~\cite{pmlr-adafactor}. The network is trained for 100 epochs. %
The dimension of the first message passing projection is set to $D=256$ and $2D$ for the second round. Both use $4$ attention heads. For localisation, we ignore edges with a predicted distance of more than 5m, as such high distance values are not trustworthy for the localisation. 

\noindent\textbf{Evaluation Measures.} We evaluate the performance in terms of both the proximity prediction and target object localisation. For the edge proximity prediction, we report the \textit{mean Predicted Proximity Error (mPPE)}, which is the mean absolute error between the predicted distances and the ground-truth pairwise distances between the target object and the objects in the partially known scene. 
We quantify the localisation performance by the \textit{Localisation Success Rate (LSR)}, which is defined as the ratio of the number of successful localisations over the number of tests. A localisation is considered successful if the predicted position of the target object is close to a target instance within a predefined distance. Unless stated differently, the distance threshold for a success is set to 1m. We consider LSR as the \emph{main} evaluation measure for our task. 
Finally, to quantify the localisation accuracy among successful cases, we report the \textit{mean Successful Localisation Error (mSLE)}, which is the mean absolute error between the predicted target position and the ground-truth position among all successful tests.

\label{sec:dataset}

\subsection{Dataset}
\label{sec:exp:dataset}

We built a new dataset of partial 3D scenes using sequences available in ScanNet~\cite{dai2017scannet}.  ScanNet contains RGB-D sequences taken at a regular frequency with a RGB-D camera. It provides the camera pose corresponding to each captured image, as well as the point-level annotations, i.e. class and instance id, for the complete Point Cloud Data (PCD) of each reconstructed scene.

The original acquisition frequency in ScanNet is very high (30Hz), meaning that most images are similar with redundant information for the scene reconstruction. We therefore use ScanNet\_frames\_25k, a subset provided in the ScanNet benchmark\footnote{http://kaldir.vc.in.tum.de/scannet\_benchmark} with a frequency of about $1/100^{th}$ of the initial one. 
We further divide the full RGB-D sequences of each scene into smaller sub-sequences to reconstruct the partial scenes. We vary the length of the sub-sequences to reflect different levels of completeness of the reconstructed scenes.
For each sub-sequence, we integrate the RGB-D information with the camera intrinsic and extrinsic parameters to reconstruct the PCD at the resolution of 5cm using Open3D~\cite{open3d}. %
The annotation for each point in the partial PCD is obtained by looking for the corresponding closest point in the complete PCD scene provided by ScanNet.  

\begin{figure}[t]
	\centering
        \begin{subfigure}[b]{.45\linewidth}
            \centering
              \includegraphics[width=1\linewidth]{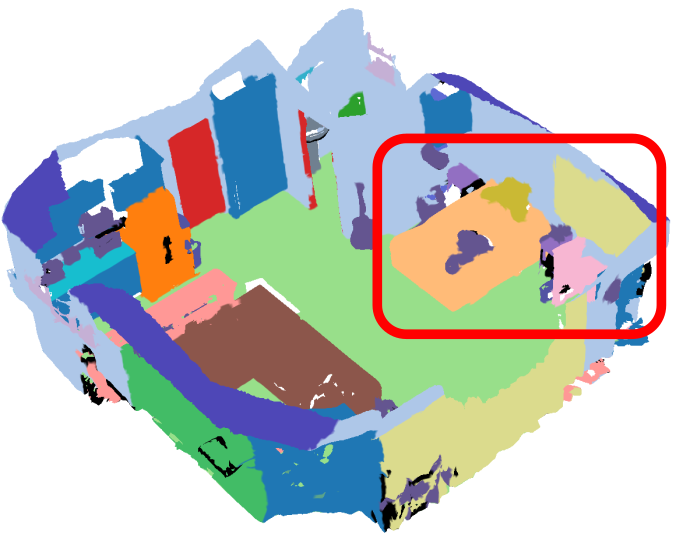}
          \caption{Complete scene}
          \label{fig:dt:A}
        \end{subfigure}%
        \begin{subfigure}[b]{.55\linewidth}
              \centering
              \includegraphics[width=1\linewidth]{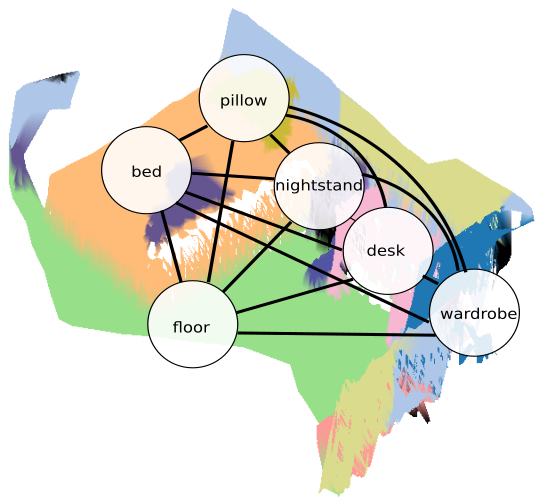}
              \caption{Partial scene}
              \label{fig:dt:B}
        \end{subfigure}%
    \vspace{-0.3cm}
    \caption{The proposed dataset with (a) the complete scene from the ScanNet dataset, and (b) our reconstructed partial scene overlaid with the Spatial Graph.}
    \label{fig:dataset_sample}
    \vspace{-0.3cm}
\end{figure}

From each partially reconstructed scene, we extract the corresponding Spatial Graph with its object nodes, i.e. the graph with only proximity edges (see Fig. \ref{fig:dataset_sample} for an example). The nodes of the graph contain the object information: e.g. the \emph{position}, defined as the centre of the bounding box containing the object, and the \emph{object class}. We consider the position of each scene object as a 2D point $(x,y)$ on the ground plane as most objects in the indoor scenes of ScanNet are located at a similar elevation.
Each node is marked as \emph{observed} if it represents an object in the partially known scene; or as \emph{unseen} if it represents the object in the unknown part of the scene, i.e. the target object to localise.

\begin{figure}[t]
          \centering
          \includegraphics[width=1\linewidth]{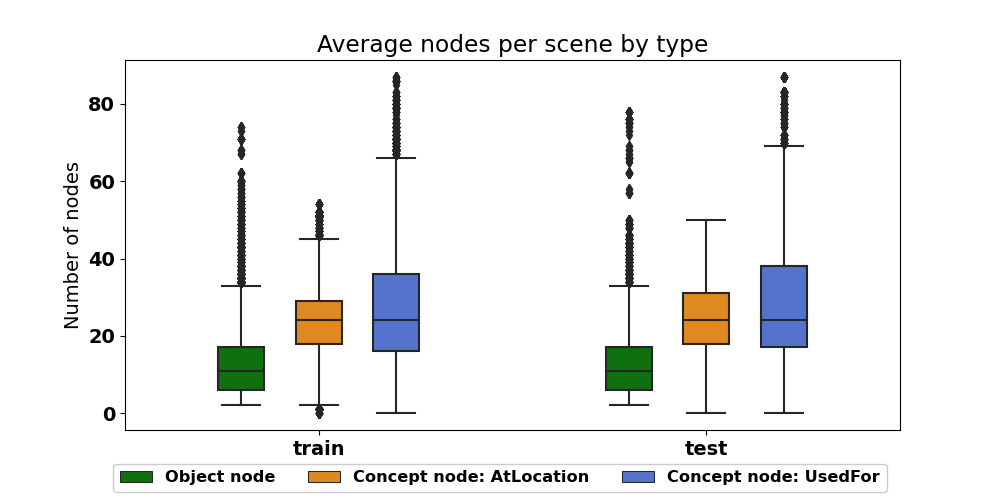}
          \caption{Average number of different types of nodes among the SCGs in the train and test split of the dataset.} %
          \label{fig:dataset_stats}
          \vspace{-0.4cm}
\end{figure}%

Moreover, we construct our SCGs %
by adding two semantic relationships \emph{AtLocation} and \emph{UsedFor}, as well as the concepts that are linked by the relationships. We extract the concepts from ConceptNet by querying each scene object using the two semantic relationships. The query returns a set of related concepts together with their corresponding weight $w$ indicating how ``safe and credible'' each related concept is to the query. We include a concept to the SCG only when it has a weight $w > 1$. Fig.~\ref{fig:dataset_stats} shows the average number of nodes linked by different types in the SCGs. On average, each SCG contains about 5 times more the concept nodes than the object nodes in the SG, demonstrating that rich commonsense knowledge is introduced within the SCG. The outliers in the boxplot visualisation are introduced by uncommon room types with a large amount of objects, e.g. libraries with several books. More statistics regarding our dataset can be found in the Supplementary Material.

Finally, we divide the dataset 
into training, validation and test sets. While we have access to the ScanNet training and validation data (1201 and 312 scenes respectively), we do not have access to their test data. To address this, we use ScanNet's validation sequences as our testing set, while 
we randomly sample a subset of scenes from the training set as the validation. By splitting ScanNet's sequences into partial reconstruction,  we have $24896$ partial scenes with $19461$ partial scenes to be used for training and validation, and $5435$ partial scenes for testing; where each partial scene has its corresponding SCG.

\subsection{Experimental Comparisons}
\label{sec:exp_val}
We validate %
\mname~by comparing its performance on our new dataset against a set on baselines and state-of-the-art methods for layout prediction.
All the baselines follow the two-staged pipeline by first predicting the pairwise distances and then estimating the position with the localisation module. We summarise below all the evaluated approaches.
\begin{itemize}[noitemsep,nolistsep]
    \item \textbf{Statistics-based baselines} uses the statistics of the training set, i.e. the \textit{mean}, \textit{mode}, and \textit{median} values of the pairwise distances between the target object and the scene objects, as the predicted distance.
    
    \item \textbf{MLP} learns to predict the pairwise distance between the target object and every other observed object in the scene without considering neither the spatial nor the semantic context. The input to this model is a pair of the target object and the observed object with each object represented by a one-hot vector indicating the class. This input is passed to a MLP that predicts pairwise distances.
    
    \item \textbf{MLP w Commonsense} learns to predict the pairwise distance between the target object and every other observed object in the scene without considering the spatial context. We first use GCN to propagate the conceptnet information to object nodes, then the features are passed to a MLP that predicts pairwise distances.

    \item \textbf{LayoutTransformer}~\cite{gupta2021Layout} uses the transformer's self-attention to generate the 2D/3D layout in an auto-regressive manner. We describe the observed objects as a sequence of elements as in \cite{gupta2021Layout}, where each element contains the object class and the position $(x, y)$. We then feed the class of the target object to generate its corresponding position $(x, y)$. For a fair comparison, we retrain the model with our training set.

    \item \textbf{\textbf{GNN w\textbackslash o Commonsense}} 
    is a variant of our approach that we have implemented to test the capability of our method when used without commonsense knowledge. The input is the Spatial Graph, which is composed only by the object nodes and proximity edges. The initial node features are not word embeddings, but are learned during training via an embedding layer. We test our model with both learnable node embeddings and using pretrained node embeddings from conceptnet.

\end{itemize}

\begin{table}[t]
\centering
\caption{Methods comparison for object localisation in partial scenes. 
mPPE: mean Predicted Proximity Error. mSLE: mean Successful Localisation Error. LSR: Localisation Success Rate (the \textit{main} measure). SG: Spatial Graph. SCG: Spatial Commonsense Graph.}
\resizebox{1\linewidth}{!}{
\begin{tabular}{|c|c|c|c|c|}
\hline
Method &
  Data Structure &
  mPPE (m) $\downarrow$ &
  mSLE (m) $\downarrow$ & \textbf{LSR} $\uparrow$\\ \hline \hline
  Statistics-Mean & Pairwise  & 1.167 &  0.63 & 0.140\\ \hline 
  Statistics-Mode & Pairwise  & 1.471 &  0.63 & 0.149\\ \hline
  Statistics-Median & Pairwise & 1.205 & 0.64 & 0.164\\ \hline

  MLP & Pairwise  &  1.165 &  0.62 & 0.143\\ \hline
  MLP w Commonsense & Pairwise  &  1.090 &  0.64 & 0.163\\ \hline

  LayoutTransformer~\cite{gupta2021Layout}& List  & - & \textbf{0.59} &0.176\\ \hline
  GNN w\textbackslash o Commonsense & SG  & 0.998 & 0.61 & 0.212\\  \hline
  \mname (Ours) - Learned Emb & SCG & 0.974 & 0.61 & 0.234\\  \hline
  \mname (Ours) - Concept. Emb & SCG & \textbf{0.965} & 0.61 & \textbf{0.238}\\  \hline
\end{tabular}
}
\label{table:baselines}
\vspace{-0.3cm}
\end{table}

\noindent\textbf{Discussion.}
Table~\ref{table:baselines} reports the localisation performance measures in terms of mPPE, LSR, and mSLE, of all compared methods evaluated on our dataset comprised of partially reconstructed scenes.
We can observe that methods with only pairwise inputs, e.g. statistics-based approaches or MLP, lead to worse performance compared to methods that account for other objects present in the observed scene.
Nevertheless, introducing some semantic reasoning on top of these methods seems to improve the performances, as shown by MLP w Commonsense that can improve by 2\% its LSR compared to the standard MLP.
LayoutTransformer directly predicts the 2D position of the target object
by taking as input the list of all the observed scene objects and using the target class as the last input token. LayoutTransformer can better encode the spatial context and outperforms the statistic-based and MLP baselines. 
The graph-based methods achieve the highest performances, suggesting that for this problem a graph-based representation of the scene is more effective than a list-based one. %
Our \mname~ that use the full SCG is able to improve on all metrics w.r.t. the \textbf{GNN} without Commonsense knowledge, when using either embeddings learned during training and pretrained conceptnet embeddings, showing how the SCG can effectively be used to improve the localisation problem. The better performances when using the pretrained embeddings are probably due the fact that these embeddings are learned on a bigger set of tasks and data, and can therefore include some information that are not limited to objects and relations used in this localisation task.

\begin{figure}[t]
	\centering
        \begin{subfigure}[b]{.45\linewidth}
            \centering
              \includegraphics[width=1\linewidth]{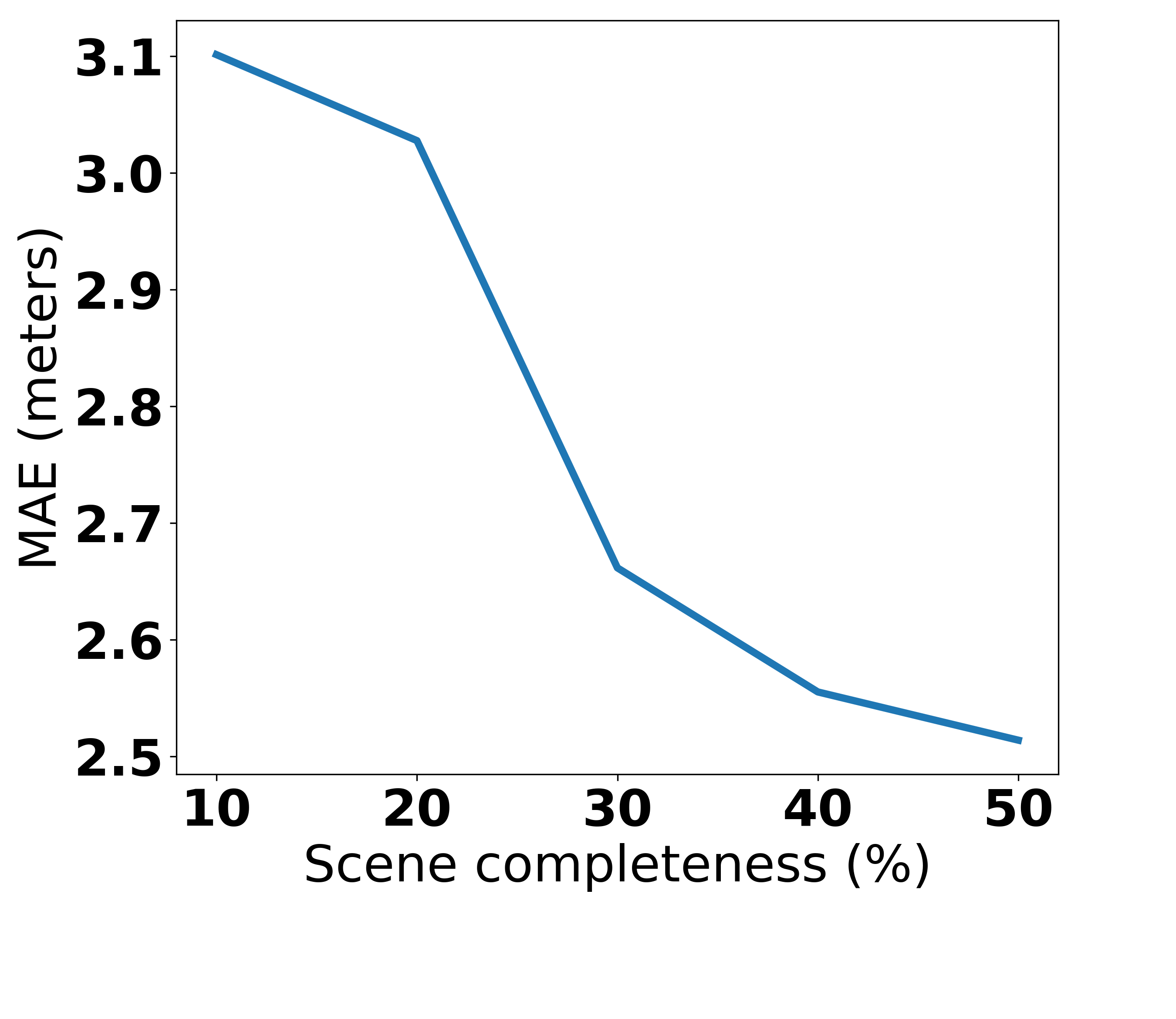}
          \caption{Localisation error}
          \label{fig:abl:A}
        \end{subfigure}%
        \begin{subfigure}[b]{.45\linewidth}
              \centering
              \includegraphics[width=1\linewidth]{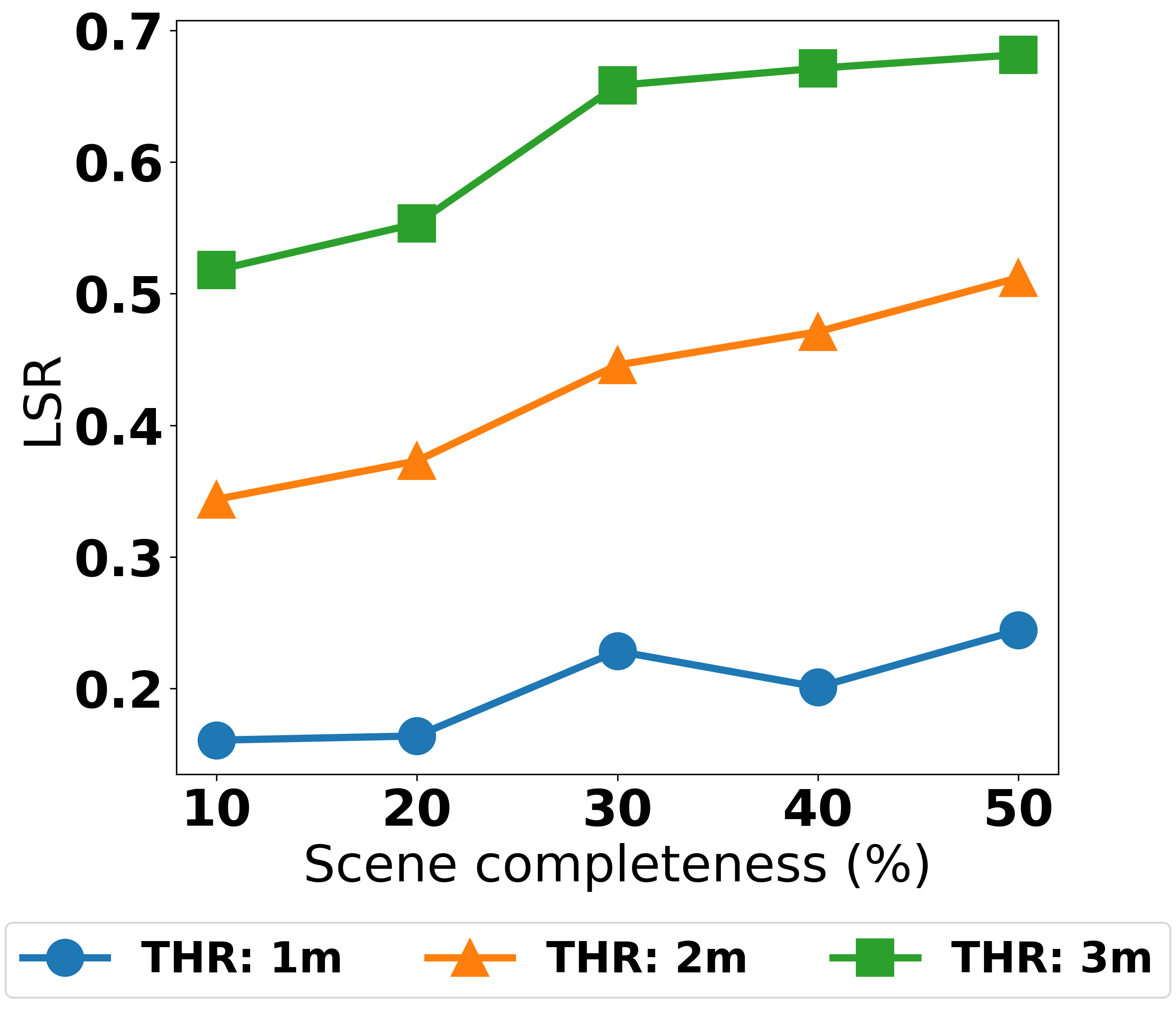}
              \caption{LSR}
              \label{fig:abl:B}
        \end{subfigure}%
    \vspace{-0.3cm}
    \caption{Localisation performance over different levels of scene completeness. (a) The localisation error in terms of MAE between the estimated target position and the ground-truth position. (b) The LSR at different threshold levels.} %
    \label{fig:abl}
    \vspace{-0.5cm}
\end{figure}

Fig.~\ref{fig:abl} shows how the completeness level of the known scene impacts the localisation performance of \mname.
Fig.~\ref{fig:abl:A} reports the mean absolute error (MAE) between the estimated position and the ground-truth position in function of the scene completeness. Note that the MAE is calculated on all the test cases including both the successful and the failed ones. 
In general, with an increasing scene completeness, \mname~can predict more accurately the position of the target object. 
Fig~\ref{fig:abl:B} presents how the LSR varies as the scene gets more complete. In general, the LSR increases when the localisation error decreases.
We report the LSR at three different threshold values, i.e. 1m, 2m, and 3m, where a larger threshold leads to a larger LSR value.

\begin{figure*}[t]
	\centering
	\includegraphics[width=0.95\linewidth]{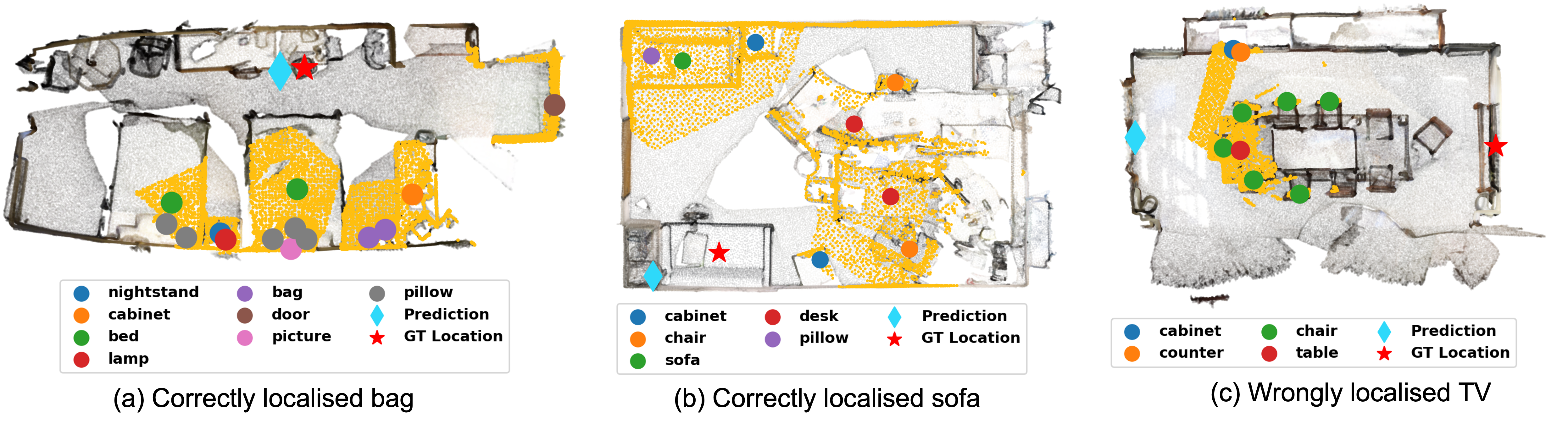}
	\vspace{-0.3cm}
    \caption{Qualitative results obtained with \mname. The partial known scene is coloured with a yellow background, while the unknown scene is indicated with grey. The  coloured circles indicate the object nodes present in the SCG. The {\color{red}red star} indicates the GT position of the target object, while the {\color{cyan} cyan diamond} indicates the predicted positions. The network is able to correctly predict the position of a bag in (a) and a sofa in (b). In the failure case of (c), the network positioned the television at the wrong side of the table. Best viewed in colour.}
    \label{fig:test}
    \vspace{-0.3cm}
\end{figure*}

\noindent{\textbf{Qualitative results.}} 
Fig~\ref{fig:test} displays some qualitative results obtained using our method \mname. Fig.~\ref{fig:test}{\color{red}~(a)} shows that the initial ``Where is my bag?'' object class example was successfully located near the area where the bag instances are. Similarly in Fig~\ref{fig:test}{\color{red}~(b)}, the position of the second sofa in the room (target object) is correctly estimated at a position opposite to the first sofa in the SCG. Interestingly, Fig.~\ref{fig:test}{\color{red}~(c)} presents a failure case when the method locates a television at the opposite side of the ground-truth television instance. Despite the estimated position is far from the real instance, the prediction is plausible due to the symmetry of the scene. We present more qualitative results in the Supplementary Material.

\subsection{Ablation study}
\label{sec:exp_ablation}
We further analyse \mname~to justify the usefulness of the commonsense relationships and the types of attention graph networks. We also investigated the impact of increasing the number of message passing layers, as well as using only the updated features when predicting the distances. 

\noindent\textbf{Which commonsense relationship is more important?} In order to better understand the effects of using different commonsense relationships, we compare \mname~against its variants in which the SCG contains: i) only \emph{Proximity} edges without commonsense relationships, ii) \emph{Proximity} edges with \emph{AtLocation} edges, iii) \emph{Proximity} edges with \emph{UsedFor} edges, and vi) \emph{Proximity} edges with \emph{AtLocation} and \emph{UsedFor} edges. 
We report the main Localisation Success Rate (LSR) measure for all variants, as well as the scene average percentage of object nodes which are linked by 0, 1, or 2 types of semantic edges, i.e. \emph{AtLocation} and \emph{UsedFor} edges.

\noindent\textbf{Discussion.} Table~\ref{tab:ablation:relation} shows that \emph{AtLocation} is more effective than \emph{UsedFor} for localising objects.
A possible reason is that using the \textit{AtLocation} edge leads to message passing among objects that are connected in the very same location, thus prioritising information  more relevant to the localisation task.
However, the best performance is obtained when the SCG can rely on all types of edges.
Moreover, most of the object nodes (~80\%) are linked to concept nodes by both \textit{AtLocation} and \textit{UsedFor} edges. This boosts the knowledge fusion much more effectively than when only one type of semantic edges are used in the SCG.
\newline

\begin{table}[t]
\caption{Impacts of different ConceptNet relationships with the proposed \mname. LSR: Localisation Success Rate.} %
\resizebox{1\linewidth}{!}{
\begin{tabular}{|c|P{0.18\linewidth}|P{0.18\linewidth}|P{0.18\linewidth}|c|}
\hline
\multirow{2}{*}{Edge Types} & \multicolumn{3}{|c|}{Obj. linked by \emph{n} semantic edges (\%)} & \multirow{2}{*}{LSR $\uparrow$} \\ \cline{2-4}
& 0 & 1 & 2 &\\ \hline \hline

Proximity  & 100 & 0 & 0 & 0.226 \\ \hline
{\it AtLocation}, Proximity & 8 & 92 & 0 & 0.233 \\\hline
{\it UsedFor}, Proximity & 19 & 81 & 0 & 0.227\\\hline
{\it AtLocation, UsedFor}, Proximity & 8 & 12 & 80 & \textbf{0.238} \\ \hline
\end{tabular}
} 
\label{tab:ablation:relation}
\vspace{-0.4cm}
\end{table}

\noindent\textbf{Which attention network is more effective?}
We examine the usefulness of the attentional network of \mname~compared to other attention modules for the localisation task.

\begin{itemize}[noitemsep,nolistsep]
\item  \textbf{No attention:} We use GINEConv~\cite{hu2019strategies} during message passing without any attention module. 

\item \textbf{Sequential GAT:}
We use GAT~\cite{velickovic2018graph} as our attentional message passing layer. As GAT cannot distinguish heterogeneous edges and cannot be used with edge features, we use it sequentially for each semantic edge: first on the \emph{AtLocation} edges, and then on the \emph{UsedFor} edges.  We then use GraphTransformer for the message passing on the proximity edges encoding the pairwise distances on the edge feature. 

\item \textbf{Sequential GATv2:} 
This method operates similarly to Sequential GAT, but employs GATv2~\cite{brody2021gatv2} for the attention layer instead of GAT.

\item \textbf{HAN}~\cite{wang2019heterogeneous}\textbf{:} 
This method
defines multiple meta-path that connect neighbouring nodes either by specific node or edge types. It employs attentional message passing sequentially by first calculating the semantic-specific node embedding and then 
updating them by another round of attentional message passing. %
With SCG we define three sets of meta neighbours, i.e. the proximity neighbours, the \emph{AtLocation} neighbours, and the \emph{UsedFor} neighbours connected by the specific edges.

\end{itemize}

\noindent\textbf{Discussion.}
As shown in Table~\ref{tab:ablation:attention}, different attention modules can produce results that vary greatly in terms of LSR.
Among all, HAN achieves the worst performance.
Sequential GAT and Sequential GATv2 were also not as effective as 
 \mname. This could be explained by a failure to integrate semantic and spatial information into the object node representation, as the semantic edges and the spatial context are aggregated separately, in a sequential manner.
In contrast, \mname~performs simultaneous message passing on all the edge types, leading to the best localisation accuracy.

\begin{table}[t]
\centering
\caption{Impacts of different attentional networks for the object localisation task on our SCG. LSR: Localisation Success Rate.}
\resizebox{0.7\linewidth}{!}{
\begin{tabularx}{1\linewidth}{|C|C|c|}
\hline
Attentional Network & Propagation mode &LSR $\uparrow$ \\ 
 \hline \hline
No attention &  - & 0.207 \\ \hline
 GAT \cite{velickovic2018graph} & Sequential &0.212 \\ \hline
 GATv2 \cite{brody2021gatv2}& Sequential & 0.206 \\ \hline
HAN \cite{wang2019heterogeneous}& Sequential &0.205 \\ \hline 
\mname & Simultaneous &\textbf{0.238}\\ \hline %
\end{tabularx}
}
\label{tab:ablation:attention}
\vspace{-0.5cm}
\end{table}

\noindent\textbf{Do the number of message passing layers and the final node concatenation of \mname~make a difference?} We examine a set of variants of our \mname~with between 1 to 4 message passing layers. Table~\ref{tab:ablation:msgpassing} 
shows how using two message passing layers leads to the best performances: using  a single layer leads to the worst results, and using more than two fails to further improve the performances.
This happens because of the over-smoothing problem~\cite{chen-oversmoothing-2020, Oono2020Graph}, where after multiple message passing rounds, the embeddings for different nodes are indistinguishable from one another. 

Given the best layer number, we also validate the choice of concatenating the original embedding to the aggregated \emph{contextual} ones, instead of using only the aggregated features. %
Concatenation is more advantageous with a LSR score of $0.238$ while directly using the aggregated node representation obtains a LSR of $0.224$. Concatenation allows the network to develop a better understanding of the context after message passing while still remembering the initial representation.%
\newline

\begin{table}[t]
\small
\centering
\caption{Impact of different numbers of message passing layers in our \mname. LSR: Localisation Success Rate.}
\begin{tabularx}{0.9\linewidth}{|C|c|c|c|c|}
\hline
\# Layers & 1 & 2 & 3 & 4 \\ \hline
LSR $\uparrow$ & 0.190 & \textbf{0.238} & \textbf{0.238} & 0.234 \\ \hline
\end{tabularx}
\label{tab:ablation:msgpassing}
\vspace{-0.5cm}
\end{table}
\vspace{-0.5cm}
\section{Discussion}
\label{sec:conclusion}
\noindent\textbf{Conclusions.} We addressed the new problem of object localisation given a partial 3D scan of a scene. We proposed a novel scene graph model, the commonsense spatial graph, by augmenting a spatial graph with rich commonsense knowledge to improve the spatial inference. With such a graph formulation, we proposed a two-stage solution for unseen object localisation. We first predict the pairwise distances between the target node and the other object nodes using the graph-based Proximity Prediction Network, and then estimate the target object's position via circular intersection. We benchmarked our proposed method and baselines on a new dataset composed of partially reconstructed indoor scenes, and showed how our solution achieved the best localisation performance w.r.t. the other compared approaches. As future work, we will investigate the applicability of our approach to large-scale outdoor scenarios in wider geographical areas, e.g. a city.

\noindent\textbf{Limitations.}
The proposed localisation pipeline is not trainable end-to-end, as we enforce supervision on the intermediate information of the pairwise object distances rather than on the target object position. This choice allows the model to be reference-free, resulting in a better generalisation. Applying end-to-end supervision on the target position might lead to a more accurate localisation, but it is challenging to achieve without loosing the ability of generalisation.
\noindent\textbf{Broader impacts.}
Our dataset is built on top of ScanNet, featuring static indoor scenes without the involvement of any human subjects. The dataset and the proposed scene graph formulation can facilitate and motivate further research regarding scene understanding.

\noindent{}\textbf{Acknowledgements} This project has received funding from the European Union’s Horizon 2020 research and innovation programme under grant agreement No 870743. This work is partially supported by the Italian MIUR through PRIN 2017 - Project Grant 20172BH297: I-MALL - improving the customer experience in stores by intelligent computer vision, and by the project of the Italian Ministry of Education, Universities and Research (MIUR) ”Dipartimenti di Eccellenza 2018-2022.

{\small
\bibliographystyle{ieee_fullname}
\bibliography{egbib}
}

\clearpage
\appendix
\section{Introduction}
In this document, we present additional and complementary details as referred in the main paper. In particular, we detail the statistics about the proposed dataset of partial scenes in Section~\ref{sec:suppl_dataset}, and more qualitative results for our localisation approach in Section~\ref{sec:suppl_qualitative}. Finally, we discuss the potential societal impact of our research in Section~\ref{sec:suppl_ethical}.

\section{Dataset}
\FloatBarrier
\label{sec:suppl_dataset}
We provide statistics regarding the geometric arrangement of the objects, such as the number of nodes and their class over all scene graphs. We also present illustrative figures of the partially reconstructed scenes to demonstrate how our dataset is constructed starting from the original ScanNet dataset.

\begin{figure}[t]
    \centering
    \includegraphics[width=0.8\linewidth]{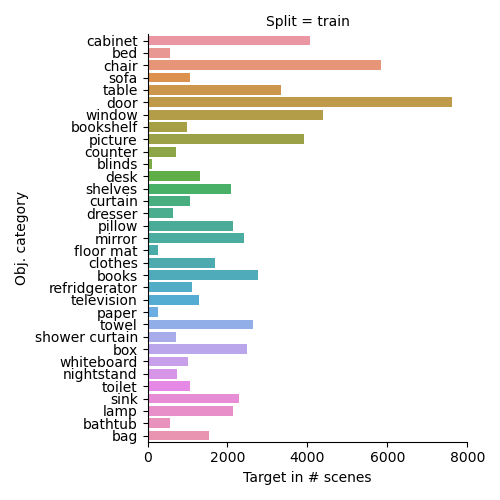}
    \\
    \includegraphics[width=0.8\linewidth]{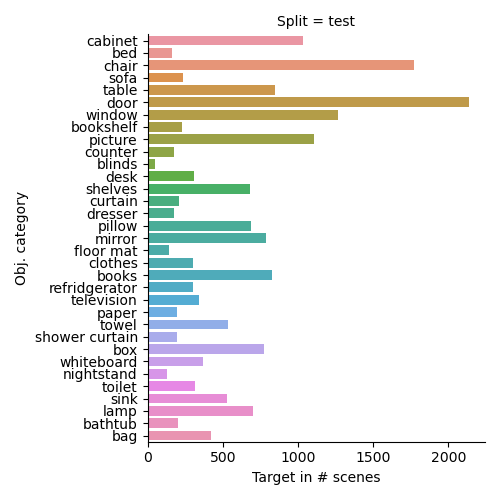}
    \caption{Numbers of scenes where an object appears as a target for the localisation. \textbf{Top:} Train set. \textbf{Bottom:} Test set.}
    \label{fig:obj_stats}
\end{figure}

\begin{figure}
    \centering
    \includegraphics[width=\linewidth]{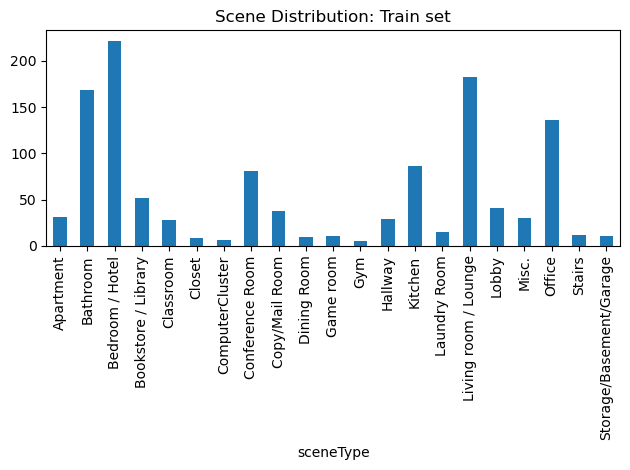}
    \caption{Room type distribution on the training set of ScanNet. The dataset mostly consists of bedrooms, bathrooms and living rooms. Some room types like closet and gym only have a few instances.}
    \label{fig:scannet_room}
\end{figure}

\subsection{Statistics}
\begin{table}[ht!]
\small
    \centering
    \caption{Statistics regarding the number of nodes in the Spatial Commonsense Graph in the train/test split.}
    \resizebox{0.8\linewidth}{!}{
    \begin{tabular}{|c||ccccc|}
    
    \hline
    & \multicolumn{5}{c|}{\# nodes}\\ 
    & Mean  & Std & Min & Max & Med  \\ \hline\hline
        Train & 59.54  & 30.06 & 3 & 177 & 55\\
        Test  & 62.17  & 32.44 & 3 & 174 & 56\\\hline
    \end{tabular}}
    \label{tab:total_stats}
\end{table}
\noindent\textbf{Total number of nodes in each scene.} Table~\ref{tab:total_stats} presents some key statistics regarding the number of nodes (both \textit{Object} nodes and \textit{Concept} nodes) in our constructed \textit{Spatial Commonsense Graph (SCG)} obtained from both the train and test split. We can observe that there is a very large variance in the number of nodes that compose the SCGs, with the smallest SCG having 3 nodes, and the largest one having about 170 nodes. This high variance is indeed a positive aspect as we can test small to wider environments populated by different numbers of objects. The varying number of objects is due to the process we use for generating each partial scene as described in section~{\color{red} 5.1} of the main paper,   i.e. by integrating RGB-D sequences of varying length. With a smaller number of RGB-D images, the reconstructed scene covers a smaller area with fewer objects, and vice versa. Another factor that contributes to the variance in the number of nodes is the addition of Concept nodes, whose number varies based on the relationship confidence queried from ConceptNet, as explained in the paper.

\noindent\textbf{Node types in each scene.} In Table~\ref{tab:nodetype_stats} we distinguish the nodes in the SCG by their type, and we present the statistics for each one obtained from both the train and test split. 
Each type of node is defined by the type of edges that the node is linked to, where Object nodes are linked by Proximity Edges, AtLocation nodes are linked by AtLocation edges, and UsedFor nodes are linked by UsedFor edges.
We observe that on average we have about 4 times more Concept nodes than the number of Object nodes. However, we notice that such a ratio does not scale to larger SCGs. The reason is that, while the Object nodes are duplicated for each object instance, the same is not true for the Concept nodes: for each Concept, only one such node exists in the graph, and multiple object nodes can be connected to it. Larger SCGs have many object nodes describing the same class, so the number of Concept Nodes does not increase linearly w.r.t. the number of object nodes. This behaviour can be observed in Fig.~\ref{fig:SCG_example}, where multiple Object nodes of the class "chair" are connected to the same Concept node. 

\begin{table*}[t!]
    \centering
    \caption{Statistics on the number of nodes in the Spatial Commonsense Graphs by node type, for both dataset partitions.}
    \resizebox{0.9\linewidth}{!}{
    \begin{tabular}{|c||ccccc|ccccc|ccccc|}
    \hline
         & \multicolumn{5}{c|}{\# Object nodes} & \multicolumn{5}{c|}{\# AtLocation nodes} & \multicolumn{5}{c|}{\# UsedFor nodes} \\
         & Mean & Std & Min & Max & Med & Mean & Std & Min & Max & Med & Mean & Std & Min & Max & Med \\ \hline \hline 
        Train & 12.77 & 8.58 & 3 & 74 & 11 & 22.10 & 9.67 & 0 & 52 & 22 & 24.67 & 14.64 & 0 & 86 & 22 \\
        Test  & 13.40 & 9.98 & 3 & 78 & 11 & 22.84 & 9.64 & 0 & 48 & 22 & 25.93 & 15.69 & 0 & 83 &22 \\ \hline
         
    \end{tabular}
    }
    \label{tab:nodetype_stats}
\end{table*}

\noindent\textbf{Distribution of target objects.} Fig.~\ref{fig:obj_stats} shows the number of partial scenes where we estimate the position of the target object category, i.e. where an instance of that object category is in the unknown part of the scene. We can see that most target objects are of categories that tend to be present in all indoor environments, e.g. doors, windows, cabinets, chairs, and pictures. This type of class imbalance is also due to the room type imbalance in the original ScanNet dataset, as shown in Fig.~\ref{fig:scannet_room} for the training split of ScanNet. Most of the reconstructed scenes are bedrooms, bathrooms, or living rooms, while other room types like closet or gym only appear a few times in the whole dataset. As such, objects that appear mostly in rooms of the minor categories will also appear less frequently as a target for our localisation task. 

\begin{table*}[t!]
\centering
\caption{Statistics of the 3D positions of objects in our dataset of partial scenes for both dataset partitions. The \emph{X,Y} plane is the floor of the room.}
\resizebox{0.9\linewidth}{!}{
\begin{tabular}{|c||ccccc|ccccc|ccccc|}
\hline
 & \multicolumn{5}{c|}{X} & \multicolumn{5}{c|}{Y} & \multicolumn{5}{c|}{Z} \\
 & Mean & Std & Min & Max & Med & Mean & Std & Min & Max & Med & Mean & Std & Min & Max & Med \\ \hline \hline
 Train & 3.63 & 2.20 & 0.01 & 15.27 & 3.34 & 3.24 & 2.06 & 0.01 & 18.06 & 3.04 & 0.86 & 0.44 & 0.03 & 4.20 & 0.76 \\
 Test  & 3.40 & 1.92 & 0.06 & 11.95 & 3.18 & 3.27 & 1.99 & 0.01 & 10.84 & 2.93 & 0.83 & 0.42 & 0.04 & 3.09 & 0.73 \\ \hline
 
\end{tabular}
\label{tab:pos_stats}
}
\end{table*}

\noindent\textbf{Geometrical arrangement of the objects.} Table~\ref{tab:pos_stats} shows the statistics on the geometrical arrangement of the objects in our dataset. While there is not much variance in the object's elevation (defined on the $Z$ axis), the variance of the object position the horizontal plane, i.e.  the $(X,Y)$ plane, is large. This indicates that in an indoor environment, the main localisation challenge lies in finding the correct position on the $(X,Y)$ plane.\\
Table~\ref{tab:dist_stats} reports the statistics on the distance between the objects in the partial scenes computed on the \emph{(X,Y)} plane. The high variance in the object position is reflected directly on the pairwise distances. 
This suggests that predicting the pairwise object distances stands for a similar difficulty as directly predicting the object position, but can better generalise to different reference systems.

\begin{table}[t!]
\small
    \centering
    \caption{Statistics on the distances between objects for both dataset partitions}
    \resizebox{0.8\linewidth}{!}{
    \begin{tabular}{|c||ccccc|}
    \hline
    & \multicolumn{5}{c|}{Pairwise distances} \\
    & Mean  & Std  & Min  & Max  & Med \\\hline\hline
        Train & 2.57  & 1.54 & 0.01 & 15.57 & 2.30\\
        Test  & 2.57  & 1.49 & 0.05 & 10.03 & 2.32\\\hline
    \end{tabular}}
    \label{tab:dist_stats}
\end{table}

Overall, these statistics show that our dataset of partial reconstructions contains very diverse scenes, with considerable variability regarding both object composition and their geometrical arrangement. Achieving a high Localisation Success Rate (LSR) on this dataset means that the method can generalise well in terms of both aspects described above.

\subsection{Examples of partial scenes}

\begin{figure*}[p]
    \centering
    \setlength{\fboxsep}{5pt}
    \setlength{\fboxrule}{0pt}
    
    \begin{tabular}{@{}c@{}|@{}c@{}@{}c@{}@{}c@{}}
    ScanNet complete PCD & \multicolumn{3}{c}{Partial reconstructions}\\ 
     \hline 
   \includegraphics[align=c,trim=0 -50pt 0 -50pt, width=0.25\linewidth]{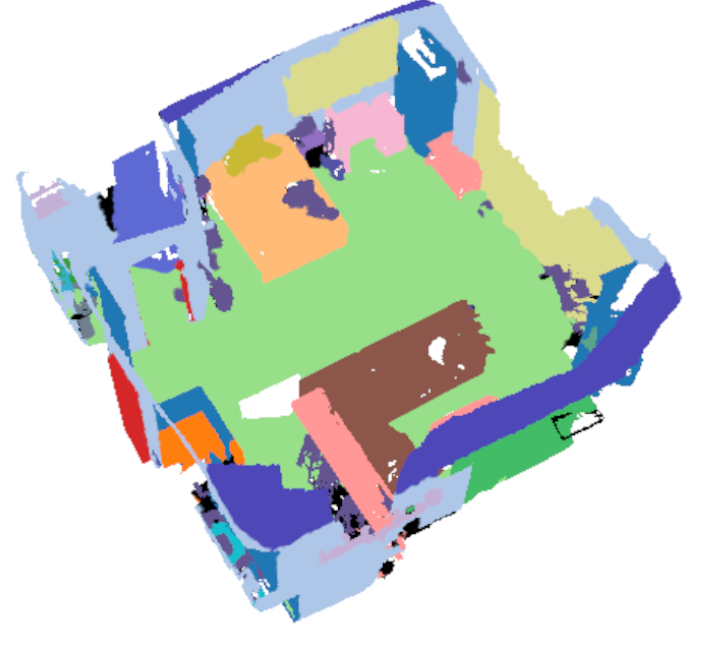} & 
    \includegraphics[align=c,trim=0 -50pt 0 -50pt,width=0.25\linewidth]{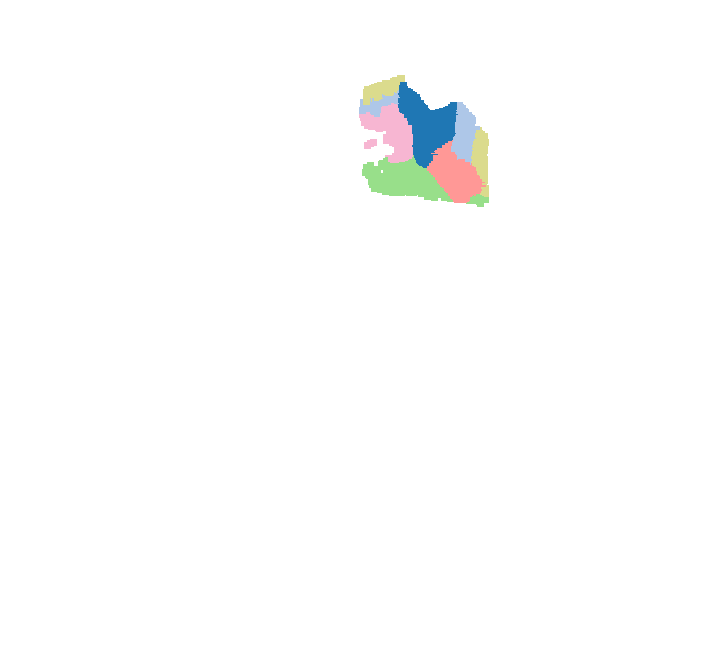} &
    \includegraphics[align=c,trim=0 -50pt 0 -50pt,width=0.25\linewidth]{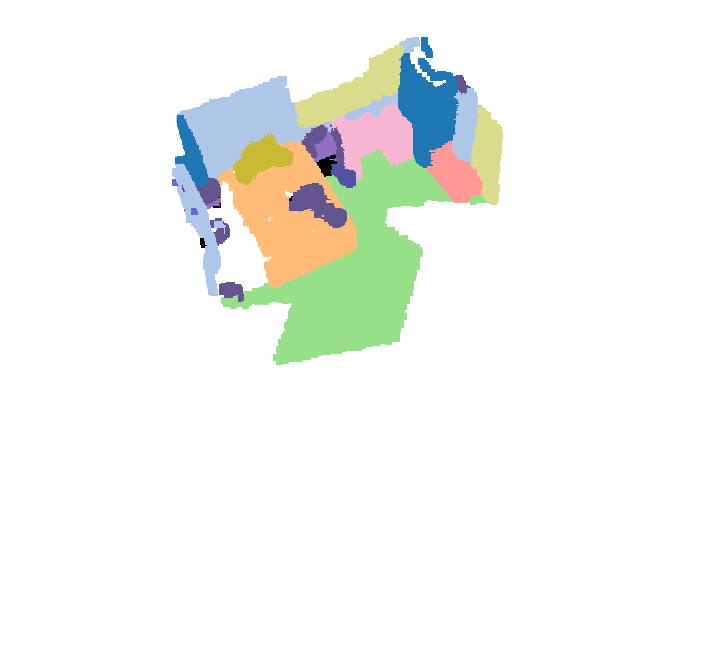} & 
    \includegraphics[align=c,trim=0 -50pt 0 -50pt,width=0.25\linewidth]{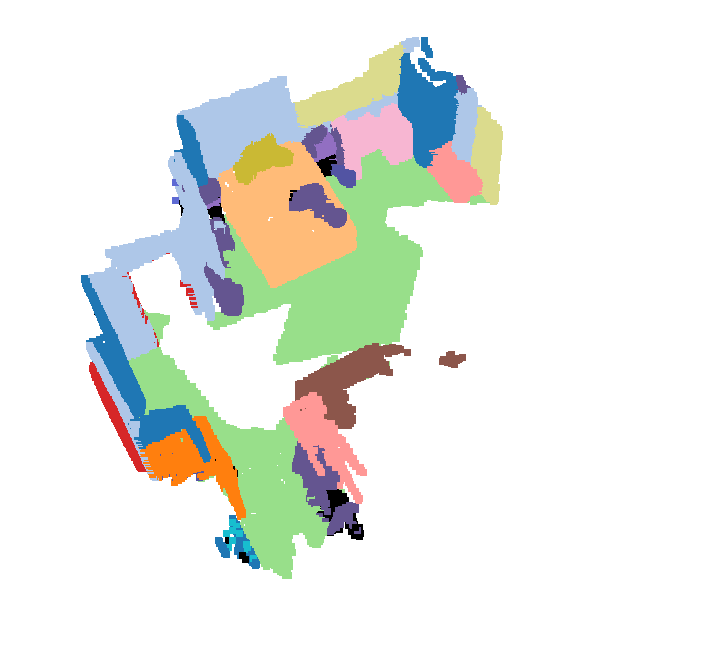} \\
    \hline
    \includegraphics[align=c,trim=0 -50pt 0 -50pt,width=0.25\linewidth]{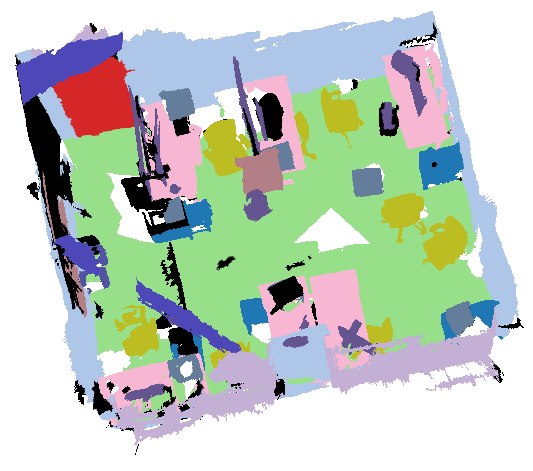} & 
    \includegraphics[align=c,trim=0 -50pt 0 -50pt,width=0.25\linewidth]{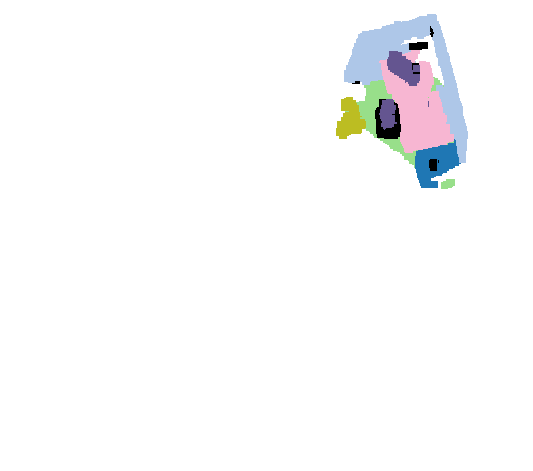} &
    \includegraphics[align=c,trim=0 -50pt 0 -50pt,width=0.25\linewidth]{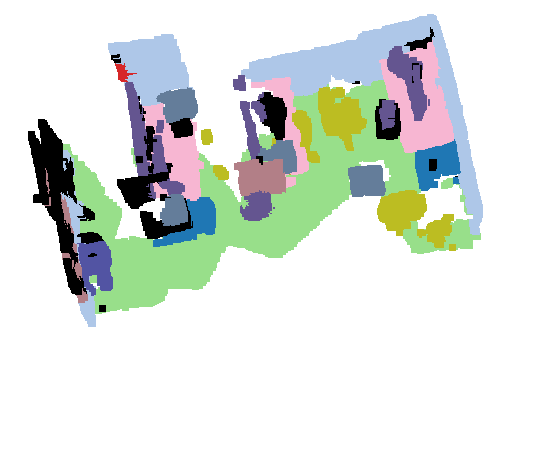} & 
    \includegraphics[align=c,trim=0 -50pt 0 -50pt,width=0.25\linewidth]{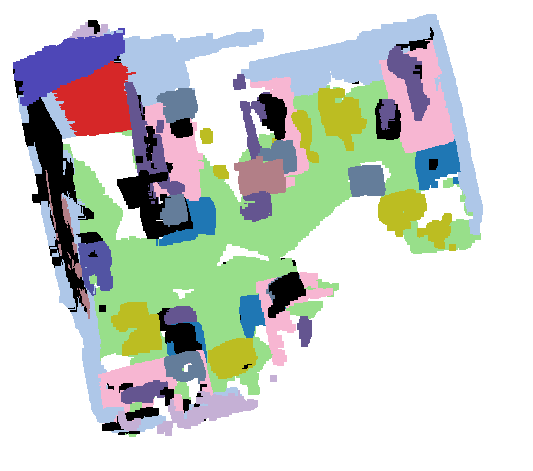}\\
    \hline
    \includegraphics[align=c,trim=0 -50pt 0 -50pt,width=0.25\linewidth]{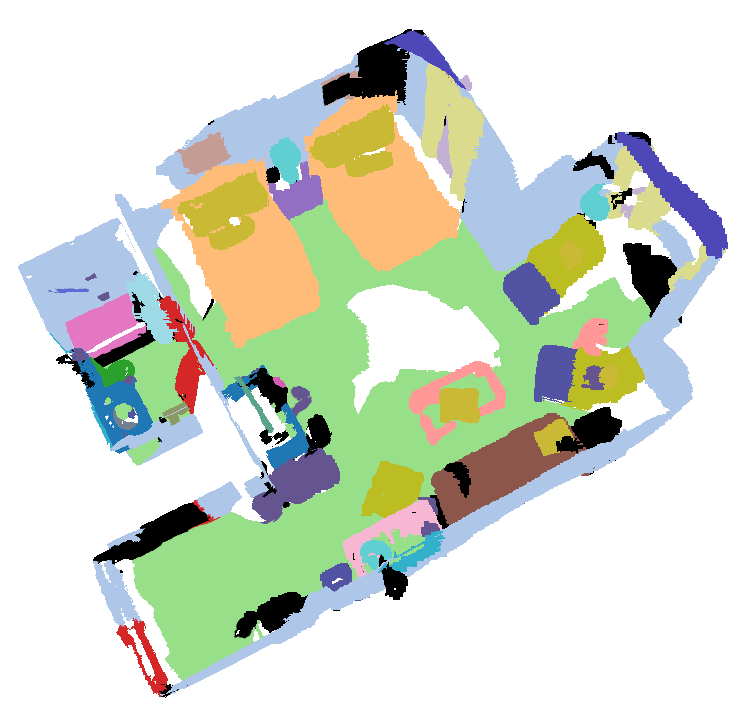} & 
    \includegraphics[align=c,trim=0 -50pt 0 -50pt,width=0.25\linewidth]{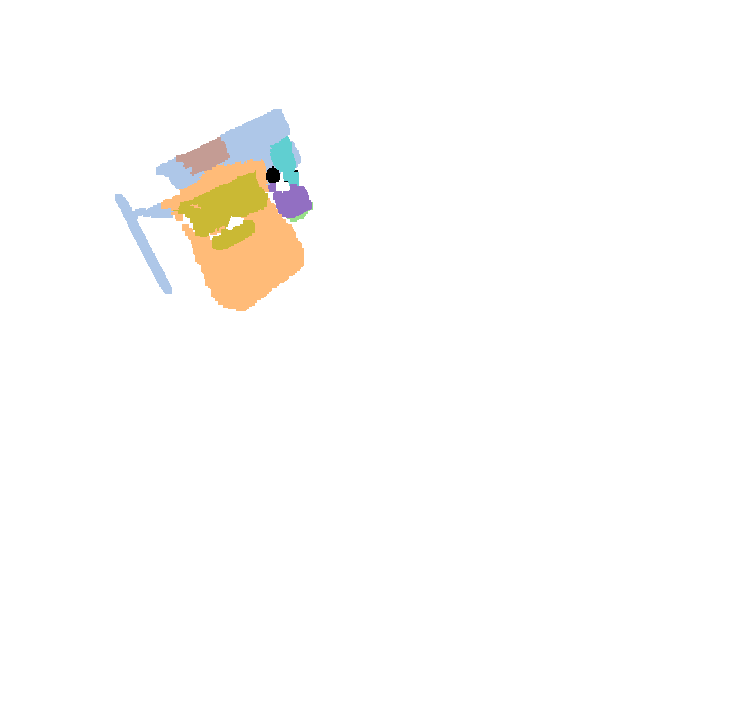} &
    \includegraphics[align=c,trim=0 -50pt 0 -50pt,width=0.25\linewidth]{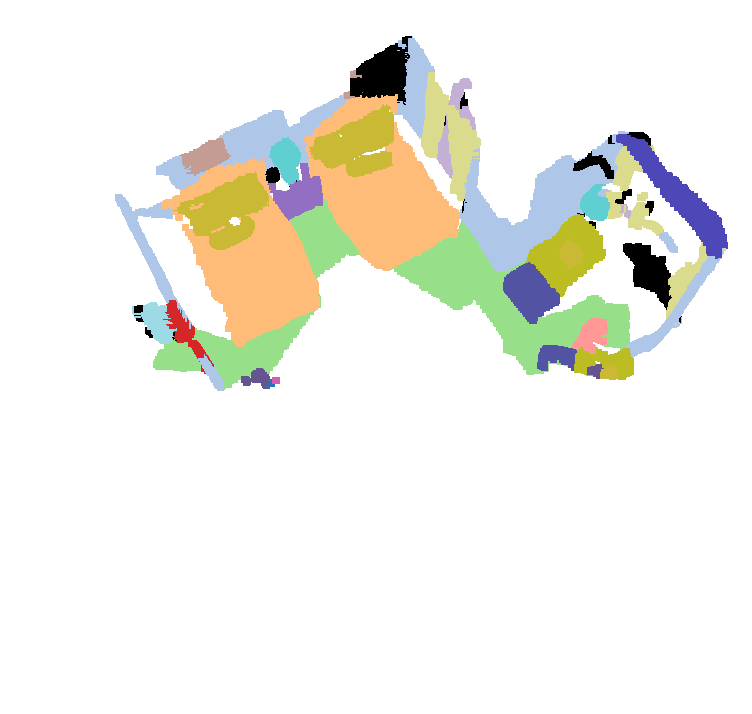} & 
    \includegraphics[align=c,trim=0 -50pt 0 -50pt,width=0.25\linewidth]{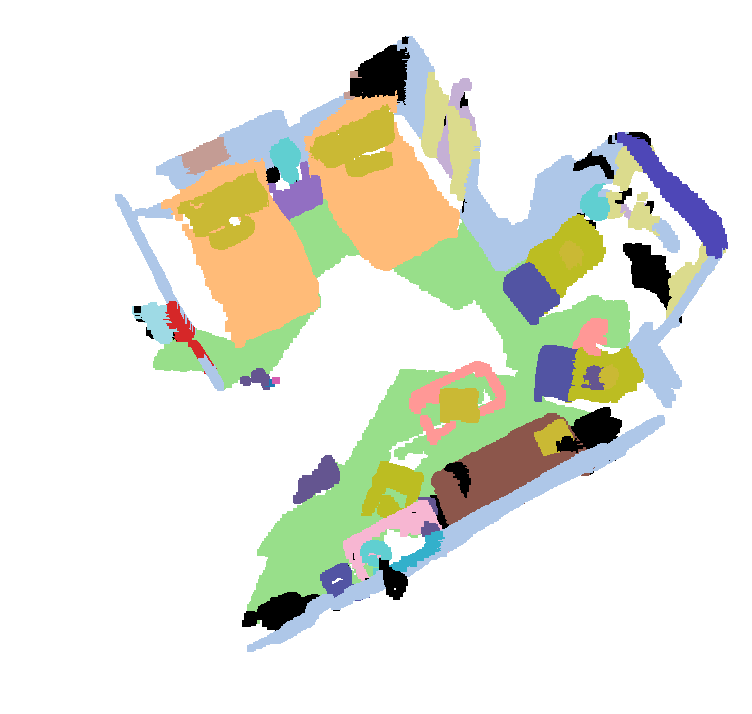}\\
    \hline
    \includegraphics[align=c,trim=0 -50pt 0 -50pt,width=0.20\linewidth]{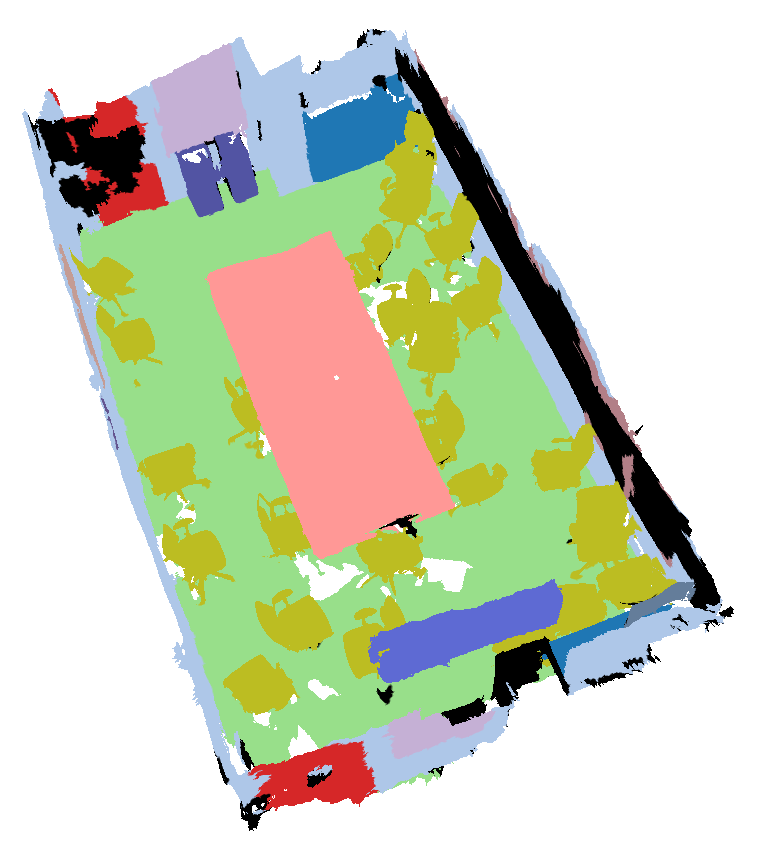} & 
    \includegraphics[align=c,trim=0 -50pt 0 -50pt,width=0.20\linewidth]{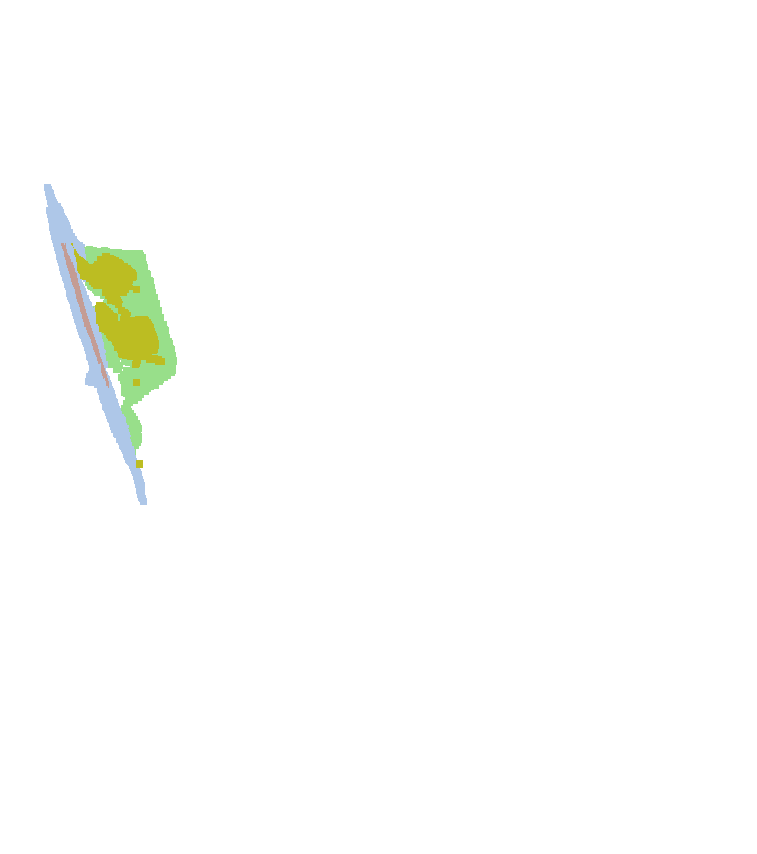} &
    \includegraphics[align=c,trim=0 -50pt 0 -50pt,width=0.20\linewidth]{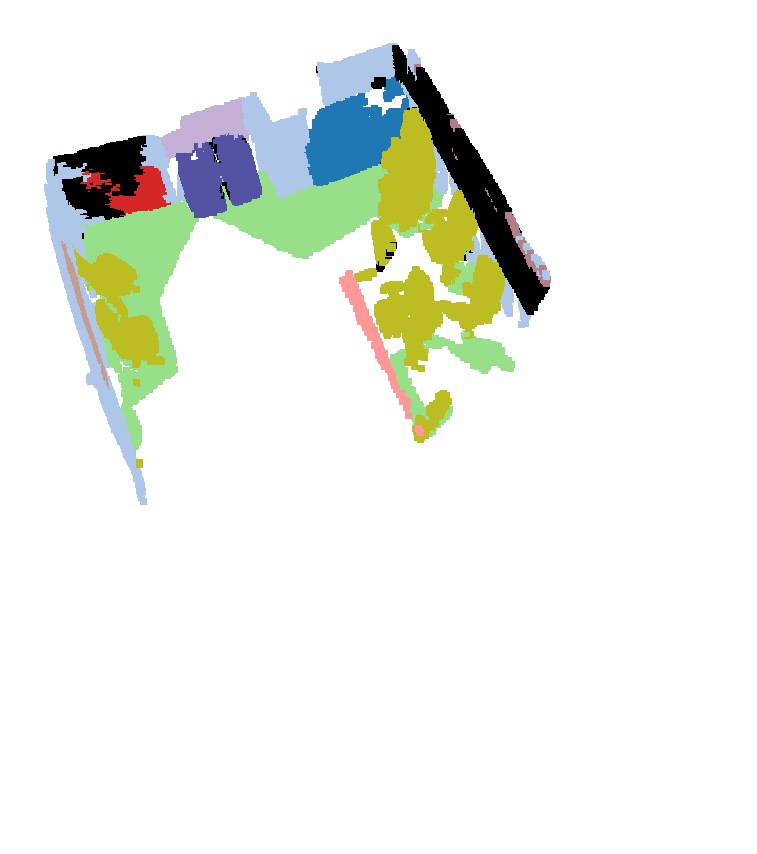} & 
    \includegraphics[align=c,trim=0 -50pt 0 -50pt,width=0.20\linewidth]{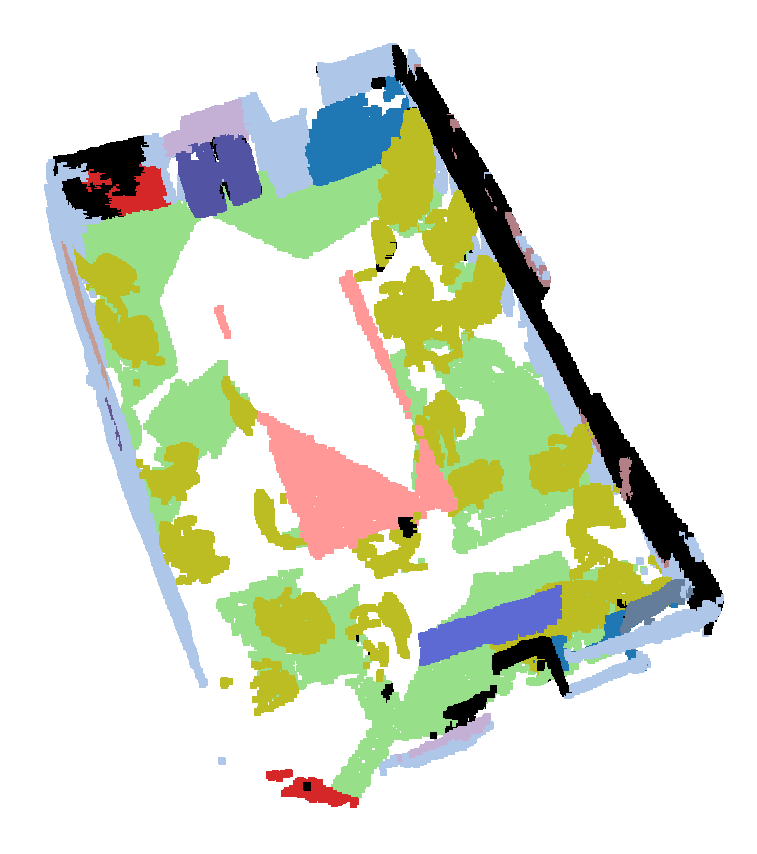}\\
    
    \
    \end{tabular}
    \caption{Examples of partially reconstructed scenes in our dataset. \textbf{Left:} The complete PCD of the room, semantically annotated with each colour indicating an object class. \textbf{Right:} Three partial reconstructions, obtained using a subset of the RGB-D sequence of increasing length (from left to right).}
    \label{fig:dataset:examples}
\end{figure*}

\noindent\textbf{Partial scenes of multiple levels of completeness.} Fig.~\ref{fig:dataset:examples} shows three examples of partially reconstructed scenes. %
To obtain the partial reconstructions, we make use of the RGB-D sequences in ScanNet which are used to reconstruct the complete scene. From the full sequence, we extract a set of subsequences of different lengths, starting from the sequence with only the first frame, to the one containing all the frames. With these sub-sequences, the extracted Point Cloud Data (PCD) tends to cover a localised area of the scene, instead of having sparse reconstruction scattered around the whole scene.
This allows us to simulate the use cases where a visually enabled device visits only a limited part of the scene with the purpose of localising a target object in the unknown part.

\begin{figure*}[p]
\centering
\includegraphics[height=0.3\textheight]{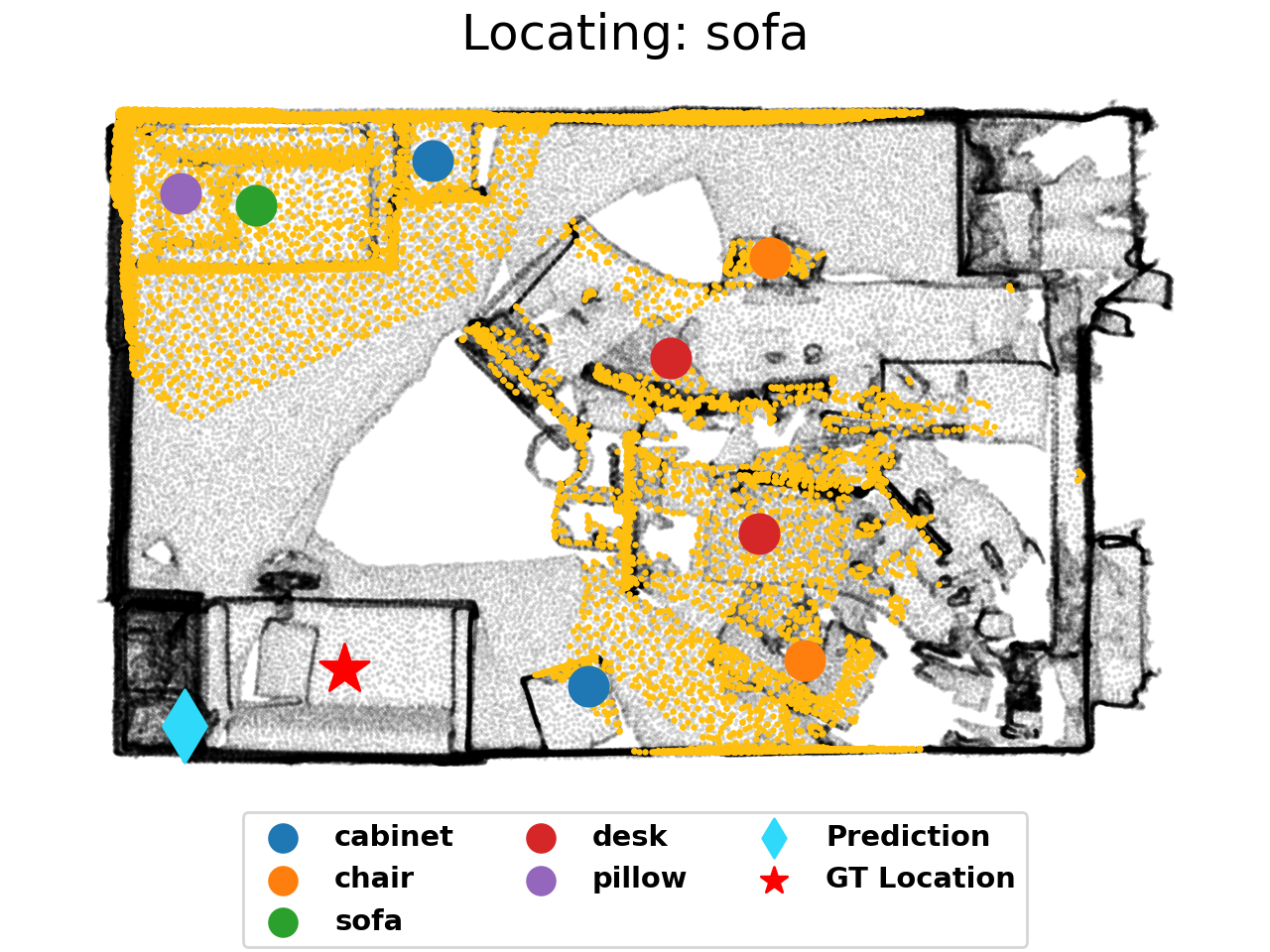}
\\
\includegraphics[height=0.6\textheight]{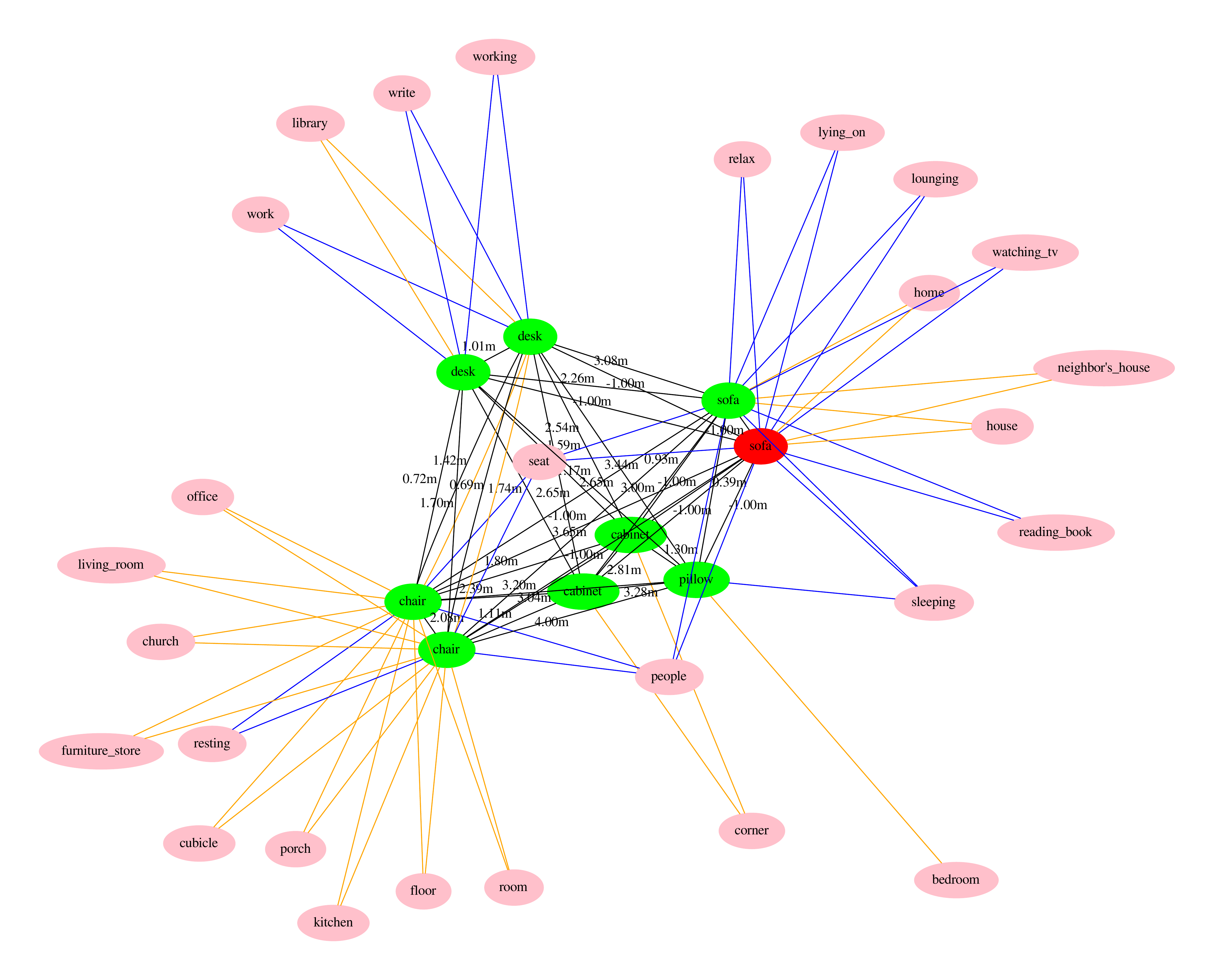}
\caption{Example of the Spatial Commonsense Graph: \textbf{Top} - Image of the partial scene with highlighted the objects in the room. \textbf{Bottom} - Spatial Commonsense Graph for the top image. The target object is represented by the {\color{red} red node}, the scene objects are the {\color{green} green nodes}, and the concept nodes have a {\color{pink} pink background}. The colour of the edge distinguish the relationship type: {\color{orange} orange} are \emph{AtLocation} edges, {\color{blue} blue} are \emph{UsedFor} edges, and {\color{black} black} are \emph{Proximity} edges.}
\label{fig:SCG_example}

\end{figure*}

\noindent\textbf{Spatial Commonsense Graph from partial scenes.} Fig.~\ref{fig:SCG_example} shows a Spatial Commonsense Graph that is related to localising a sofa. The target node representing the sofa is highlighted in red, the object nodes are highlighted in green and the concept nodes are highlighted in pink. The edges' colours describe the relationship type, with \emph{proximity} edges in black, the \emph{AtLocation} in orange, and the \emph{UsedFor} in blue. For the proximity edges we also show the pairwise distance between the objects.

We can see that some Concept nodes are connected to more than one object node, indicating a common usage or location, e.g. \emph{sleeping} for both \emph{sofa} and \emph{pillow}, or \emph{seat} for both \emph{chair} and {sofa}.

This example demonstrates how much information can be added in the scene graph composed of only 5 object nodes, by integrating commonsense knowledge with 27 Concept Nodes. 
Note that the only criterion that we apply when retrieving Concept nodes from ConceptNet is to retain nodes with a weight score above a certain threshold (relation weight~$>1$). This explains why some Concept nodes may seem not closely related to our task, e.g. sofa \emph{AtLocation} neighbour's house, chair \emph{AtLocation} furniture\_store.

\section{Qualitative results}
In this section, we show more qualitative results on the localisation with partial scenes. In particular,  we show with real examples how the Localisation Module converts from  pair-wise distance predictions to the position of the target object. Moreover, we show a comparison where we localise an object in a scene  with different levels of scene completeness. 

\begin{figure*}[p]
    \centering
    
    \begin{tabularx}{\linewidth}{CC}
    Complete Scene & Model Prediction \\ \hline
    \includegraphics[align=c,width=0.6\linewidth]{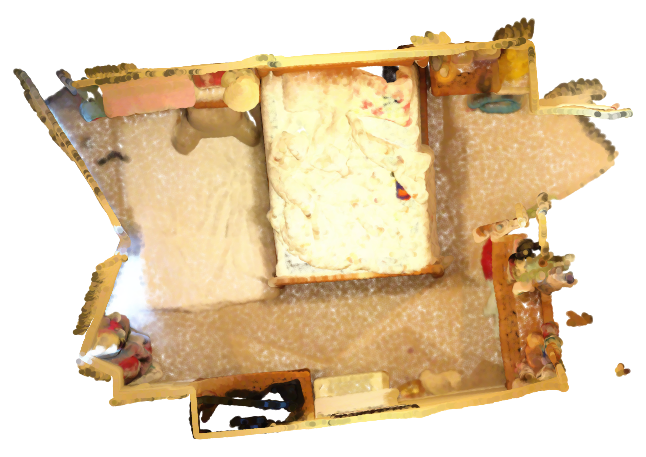} & 
    \includegraphics[align=c,width=0.6\linewidth]{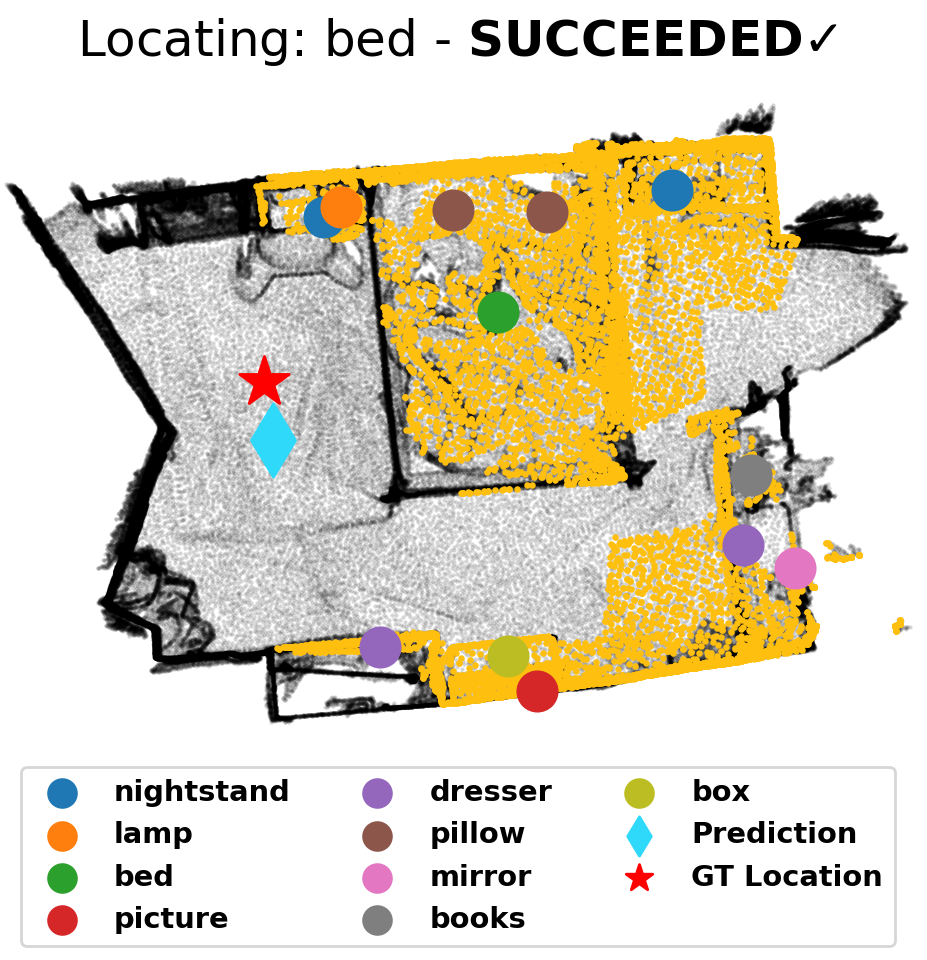} \\
    
        \includegraphics[align=c,width=0.6\linewidth]{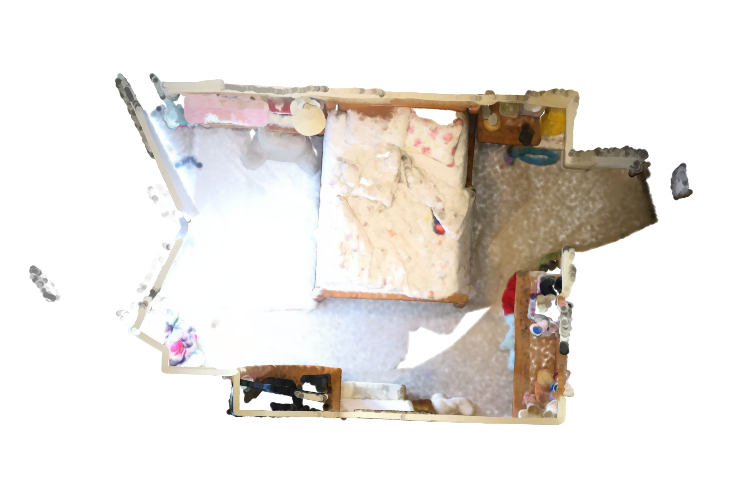} & 
    \includegraphics[align=c,width=0.6\linewidth]{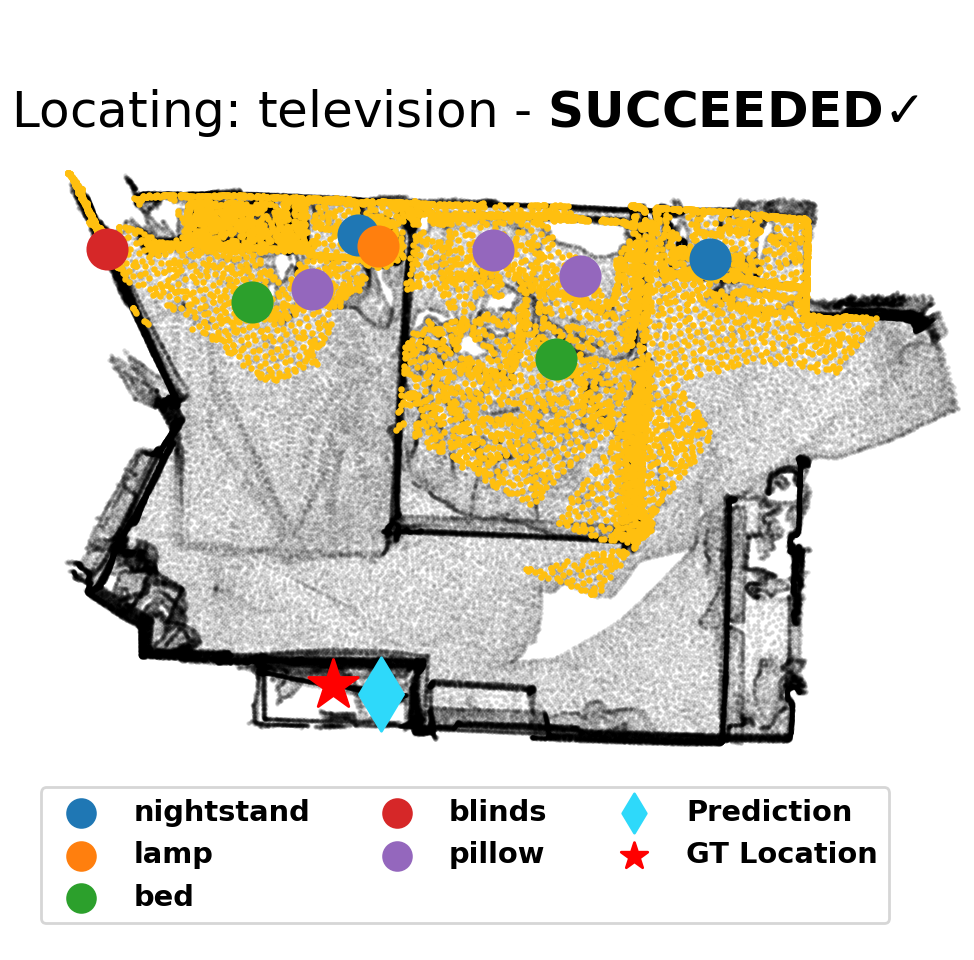} \\
    
        \includegraphics[align=c,width=0.6\linewidth]{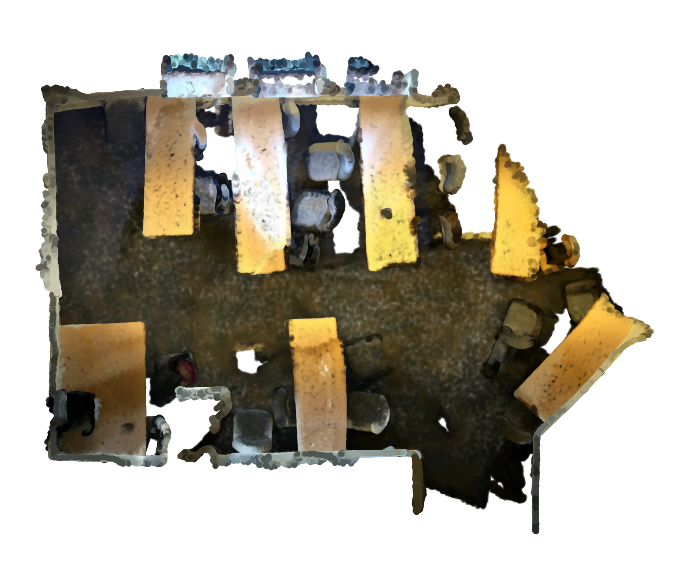} & 
    \includegraphics[align=c,width=0.6\linewidth]{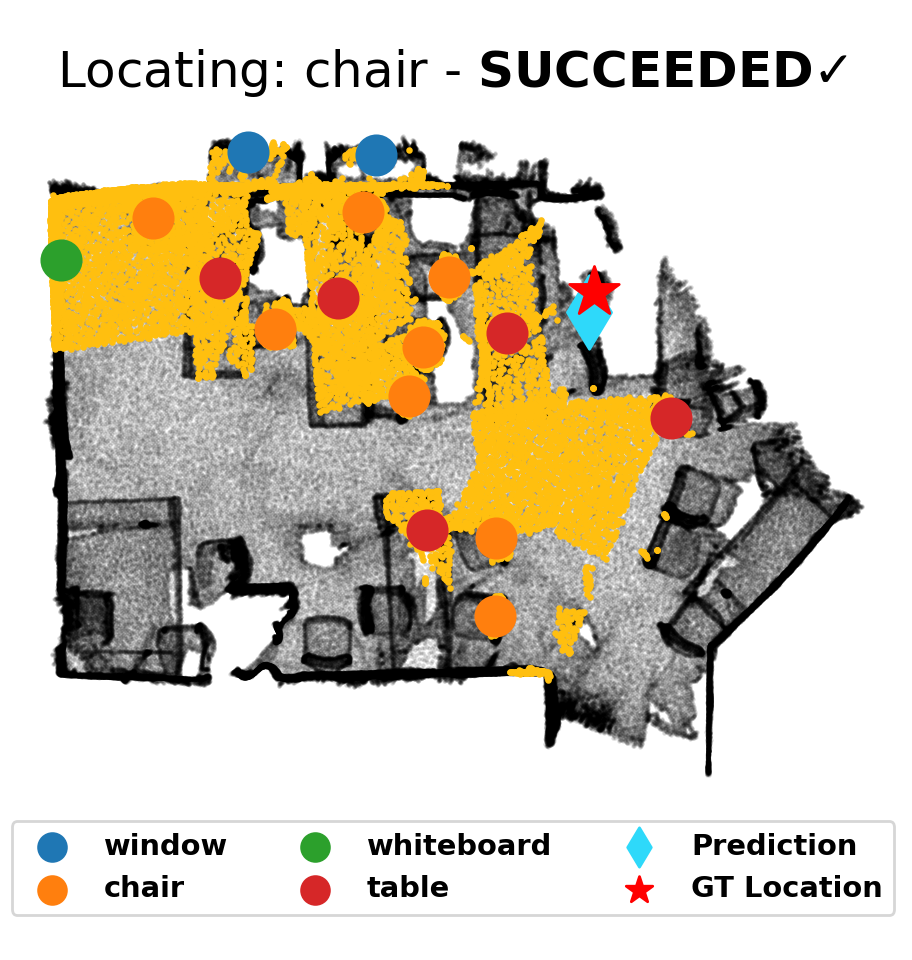} \\
    
        \includegraphics[align=c,width=0.6\linewidth]{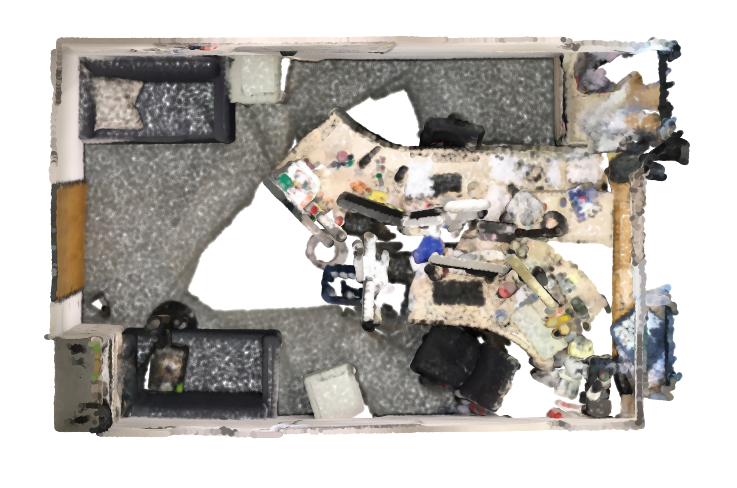} & 
    \includegraphics[align=c,width=0.6\linewidth]{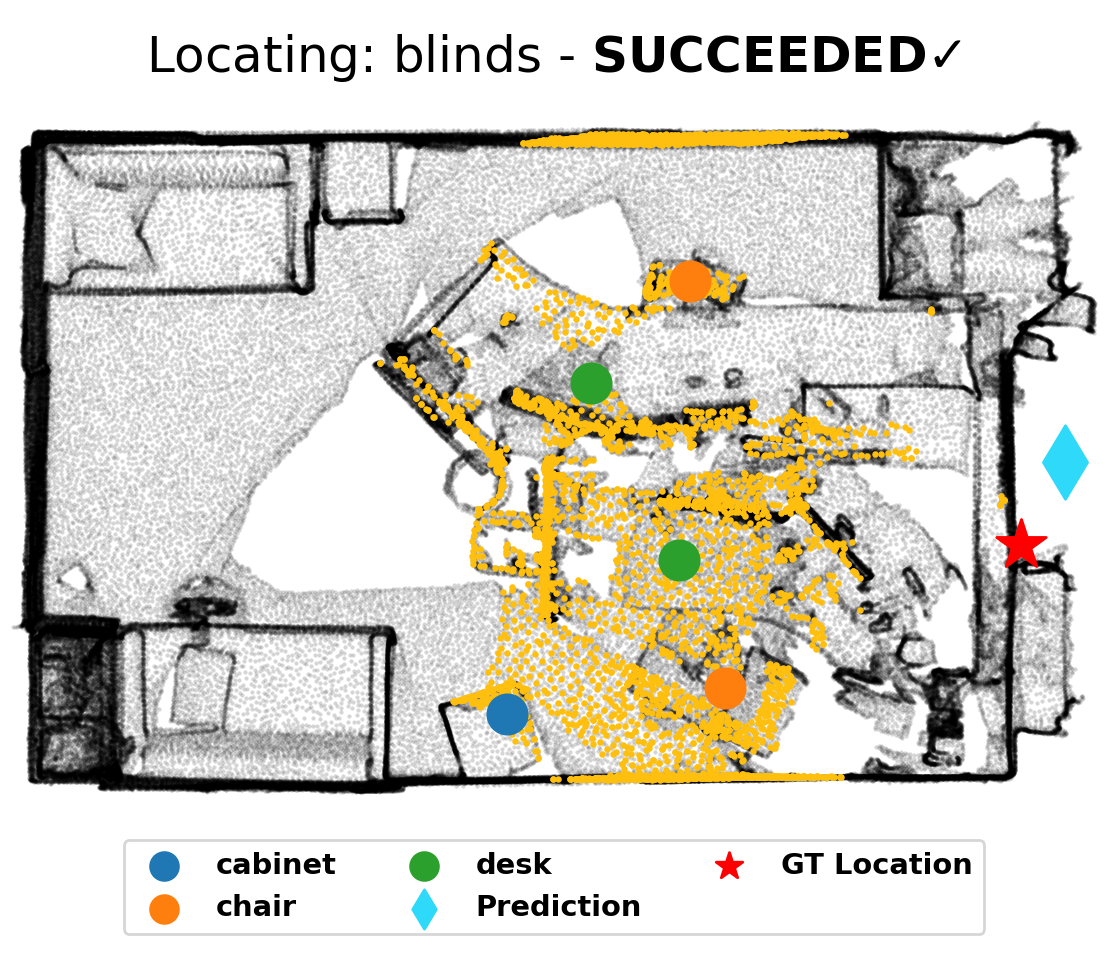}
    \end{tabularx}
    \caption{Successful localisation cases. The \textbf{left} column shows the complete scene from ScanNet. The \textbf{right} column shows the object nodes in the SCG and the position predicted by our \mnamefull\,for the target object. The yellow areas indicate the visible part of the scene. Coloured dots show the objects in the SCG. The cyan diamond indicates the predicted position and, the red start is the ground-truth position of the target instance closest to the predicted position.}
    \label{fig:qualitative}
\end{figure*}

\label{sec:suppl_qualitative}
\noindent\textbf{Additional qualitative results.}
Fig.~\ref{fig:qualitative} shows additional successful localisation with our \mnamefull. On the left, we show the coloured reconstruction of the complete scene. On the right, we display the position predicted by our method, given a partial observation of the scene, highlighted with a yellow background. For all of the four examples, our approach was able to successfully estimate the position of the target object in the unseen part of the scene. 

\begin{figure*}[p]
    \centering
    
    \begin{tabularx}{\linewidth}{CC}
    PPN Predicted edges & Localisation module \\ \hline
    \includegraphics[align=c,width=0.55\linewidth]{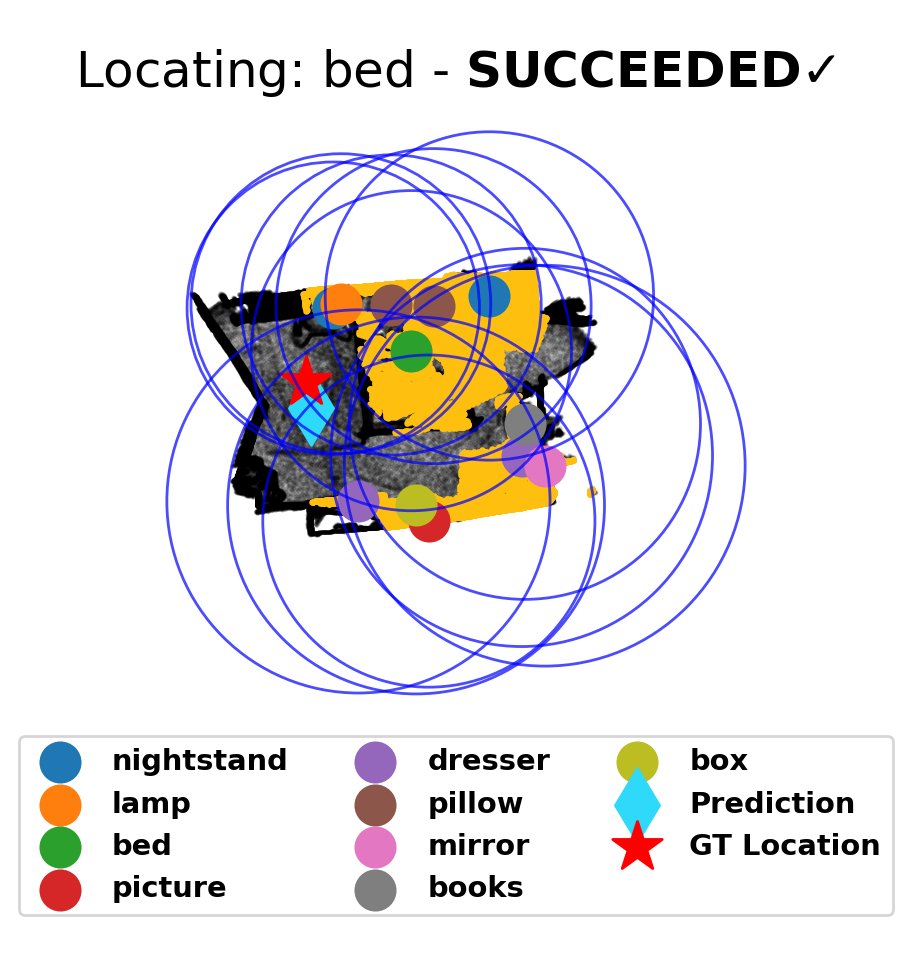} & 
    \includegraphics[align=c,width=0.55\linewidth]{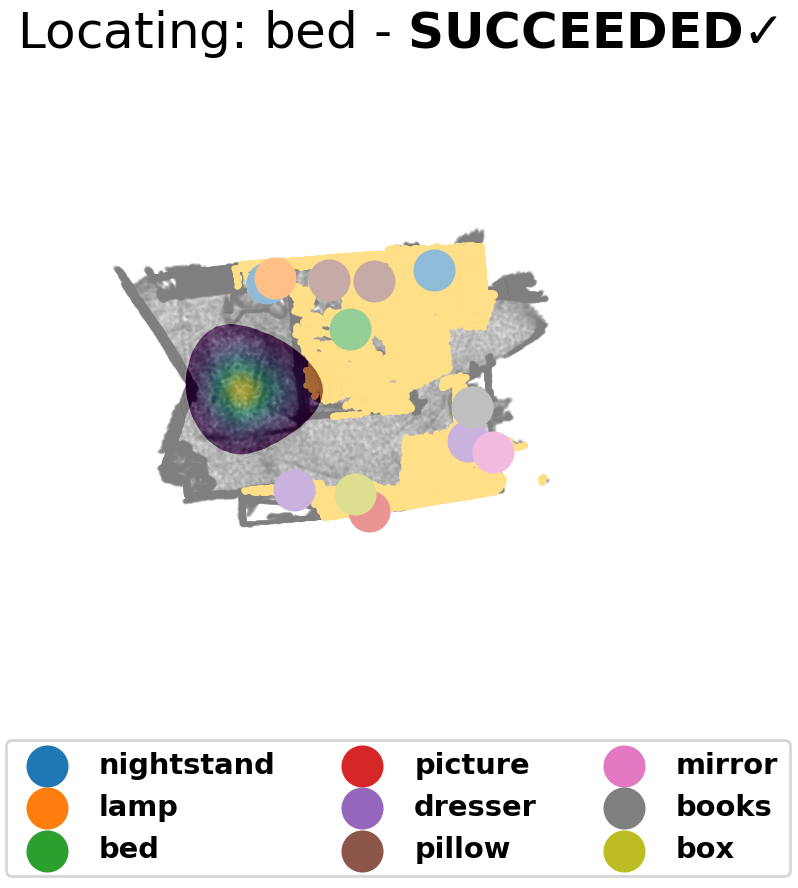} \\
    
        \includegraphics[align=c,width=0.55\linewidth]{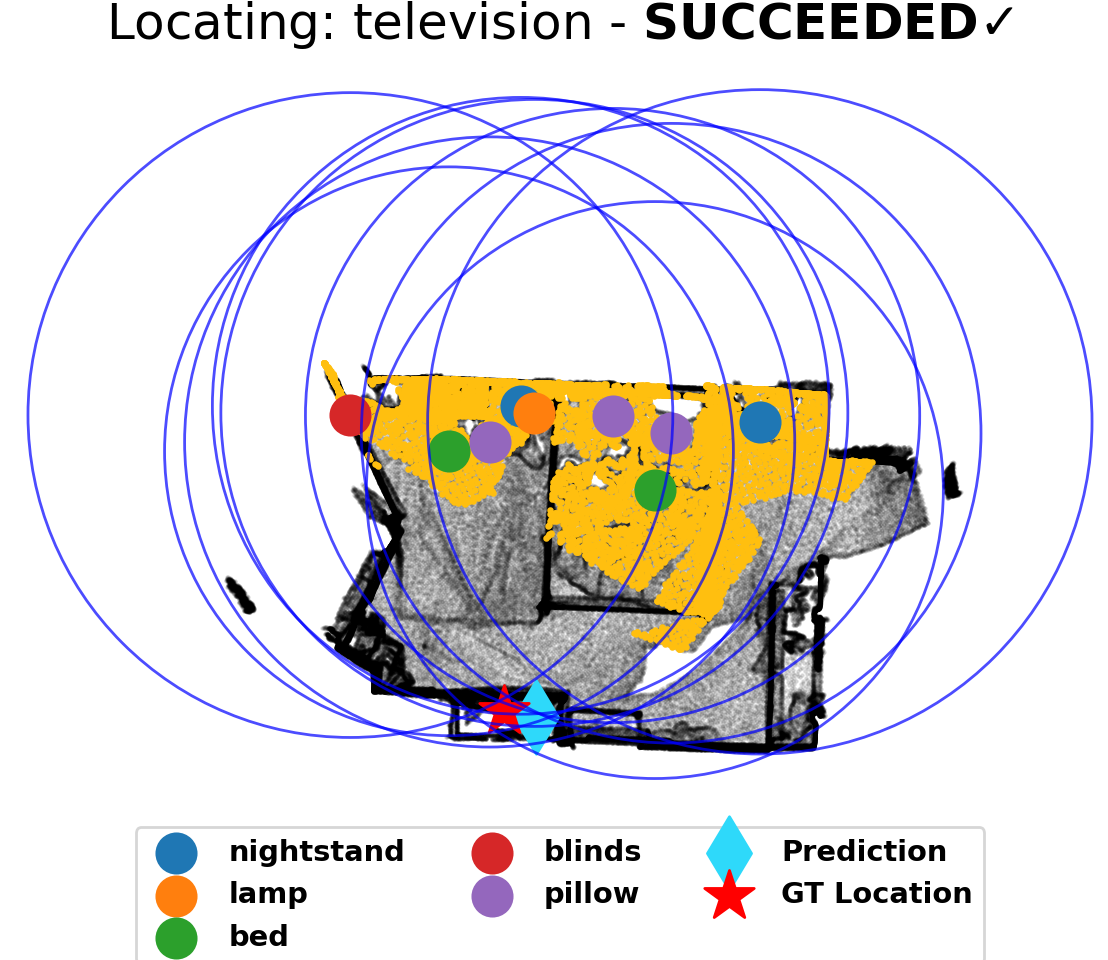} & 
    \includegraphics[align=c,width=0.55\linewidth]{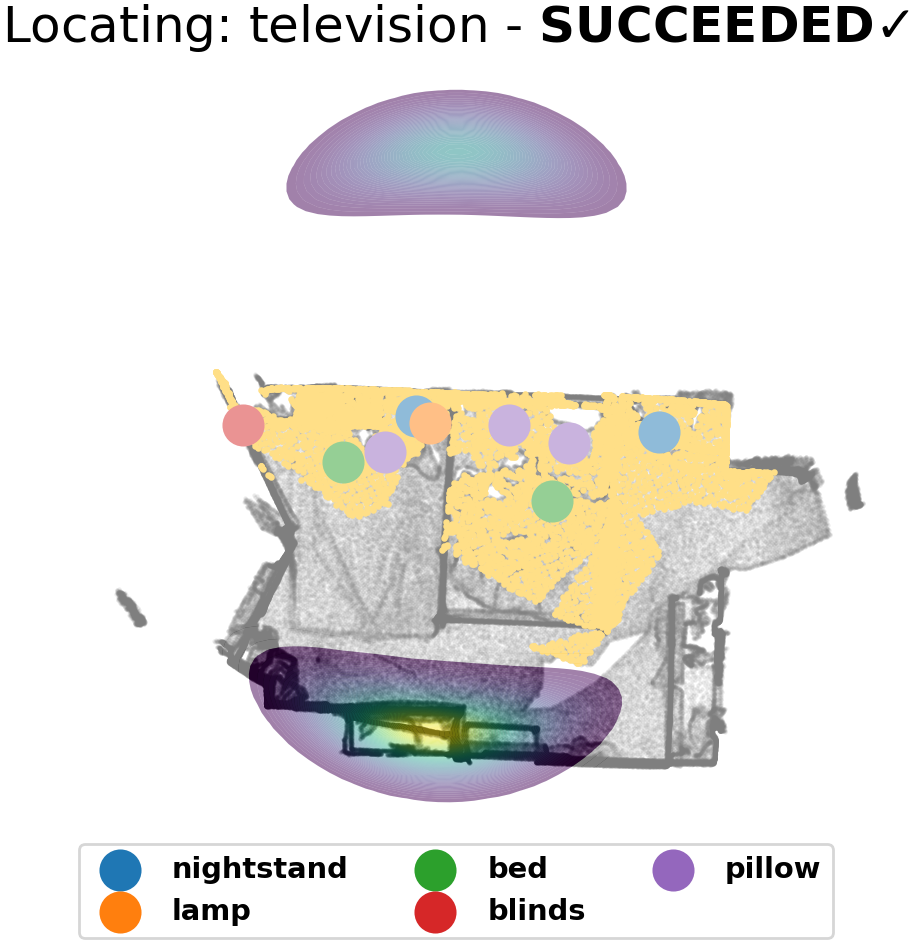} \\
    
        \includegraphics[align=c,width=0.55\linewidth]{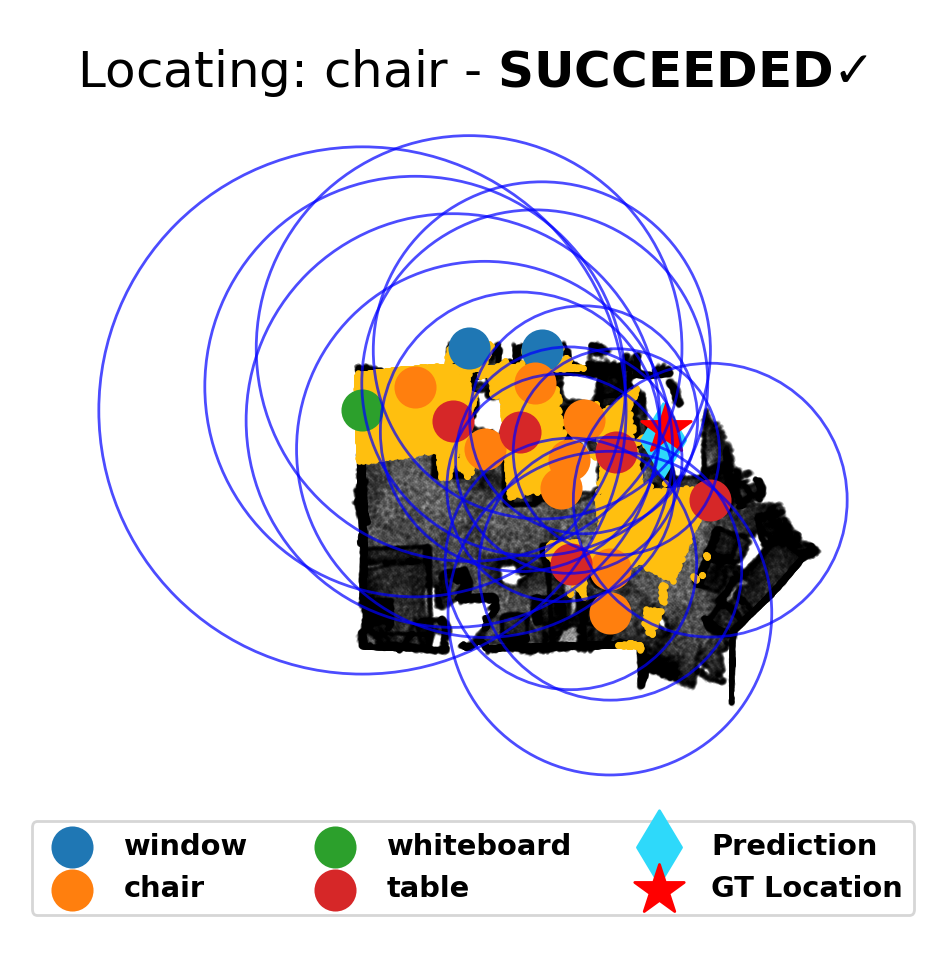} & 
    \includegraphics[align=c,width=0.55\linewidth]{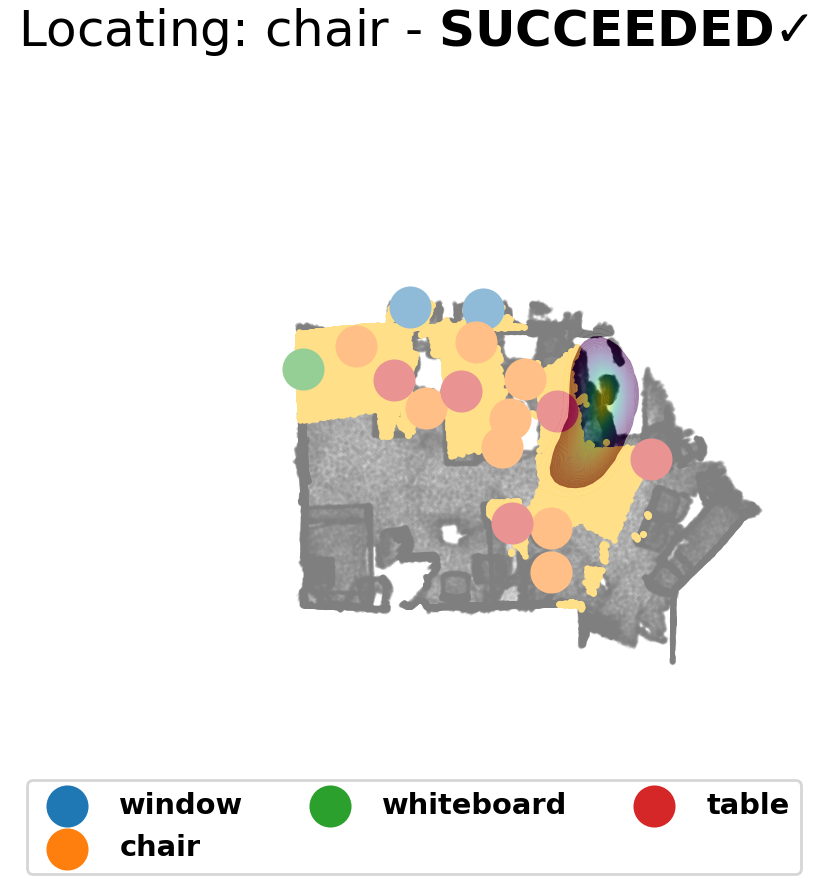} \\
    
        \includegraphics[align=c,width=0.55\linewidth]{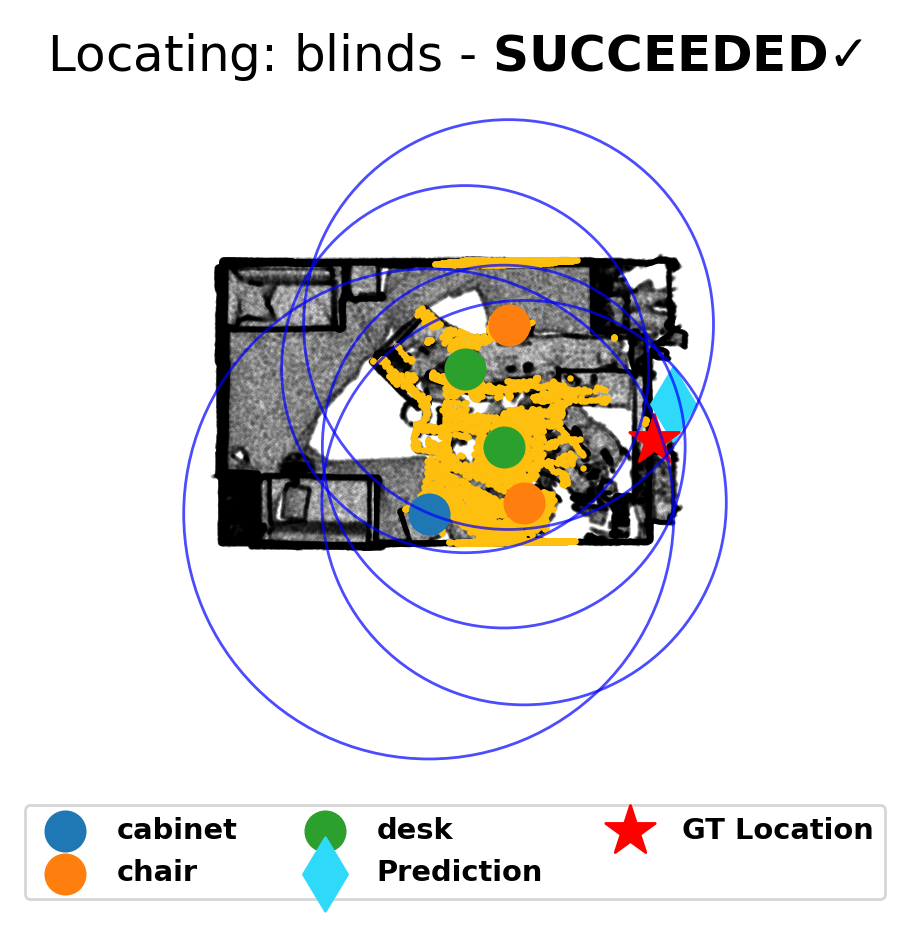} & 
    \includegraphics[align=c,width=0.55\linewidth]{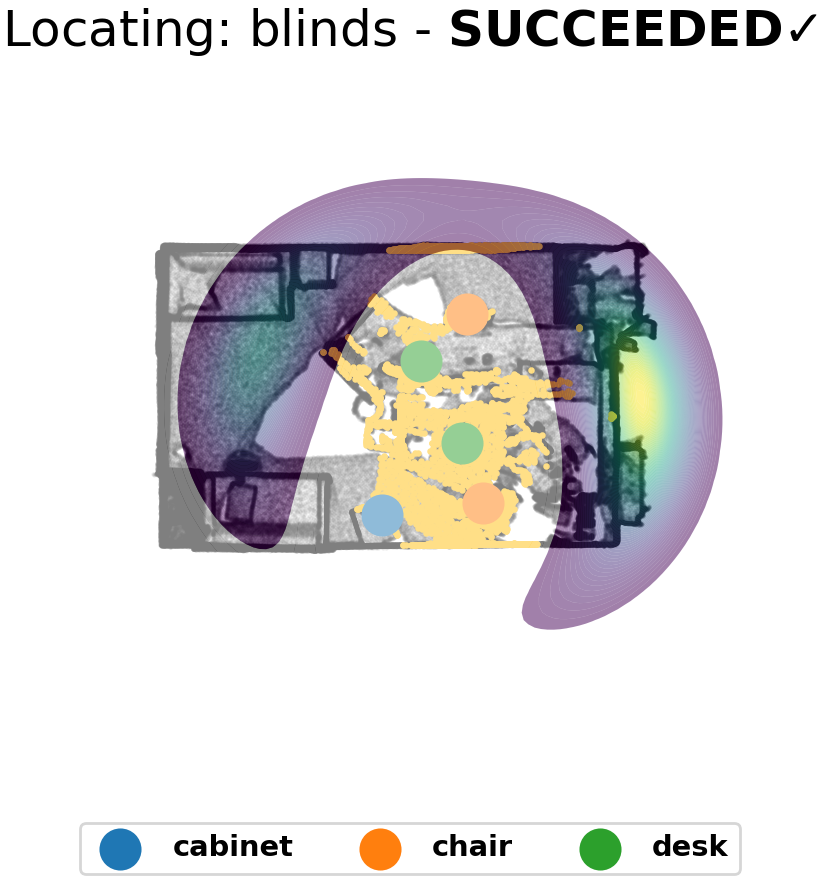}
    \end{tabularx}
    \caption{Effect of the Localisation Module, on the same examples as fig.~\ref{fig:qualitative}. The \textbf{left} column shows the edges predicted by our \emph{Proximity Prediction Network}, show in blue. The \textbf{right} column shows the cost defined in the \emph{Localisation Module}, the areas where most edges overlap have a lower cost and are displayed in yellow, while the areas with higher cost are displayed in  blue. Blank areas have a cost above the threshold set for the visualisation.}
    \label{fig:suppl:qualitative_localisation}
\end{figure*}

\noindent\textbf{Demonstration of the Localisation Module.} Fig.~\ref{fig:suppl:qualitative_localisation} demonstrates with an example how the pairwise distances predicted by our \emph{Proximity Prediction Network} are converted by our \emph{Localisation Module} to a single position in the unknown space. We define a cost function that is built with the pairwise distances of the Proximity edges, Eq.~{\color{red} 5} of the main paper. The predicted distances between the target object and each observed object in the scene are visualised as a circle centred on each observed object, as shown at the left in Fig.~\ref{fig:suppl:qualitative_localisation}, while the value of the defined cost function for all positions in the scene is visualised at the right in Fig.~\ref{fig:suppl:qualitative_localisation}. The most yellow area indicates the lowest cost and the bluest the highest cost. The position where the object is most probably located is at the position with the lowest cost, i.e. the most yellow position. From the demonstrated cases in Fig.~\ref{fig:suppl:qualitative_localisation}, we also observe that the predicted distances can be noisy with a certain degree of error. Methods with a high noise gain like Linear Least Squares, which are often used for multilateration, cannot be employed in this scenario. Differently, our approach can better tolerate erroneous distance measures. 

The cases in the second and fourth rows show the presence of multiple local minima for solving the minimisation problem. Methods that search for a local minimum, e.g. gradient descent, non-linear Least Squares, may fail to converge to the correct solution due to a bad initialisation. 
Instead, our localisation module first divides the space into a coarse grid, where the cost of each cell is calculated. The one with the lowest cost is used as the starting point to initialise the solver for the minimisation method. 
This improves the chances of converging to the global minimum.

\begin{figure*}[p]
    \centering
    \includegraphics[height=0.2\textheight]{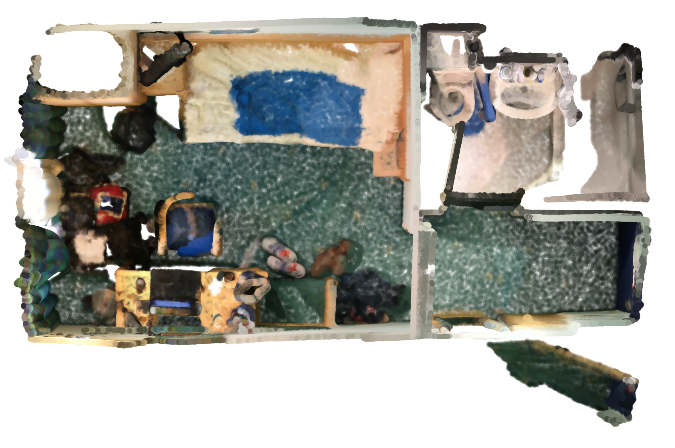} \\
    
    \includegraphics[height=0.3\textheight]{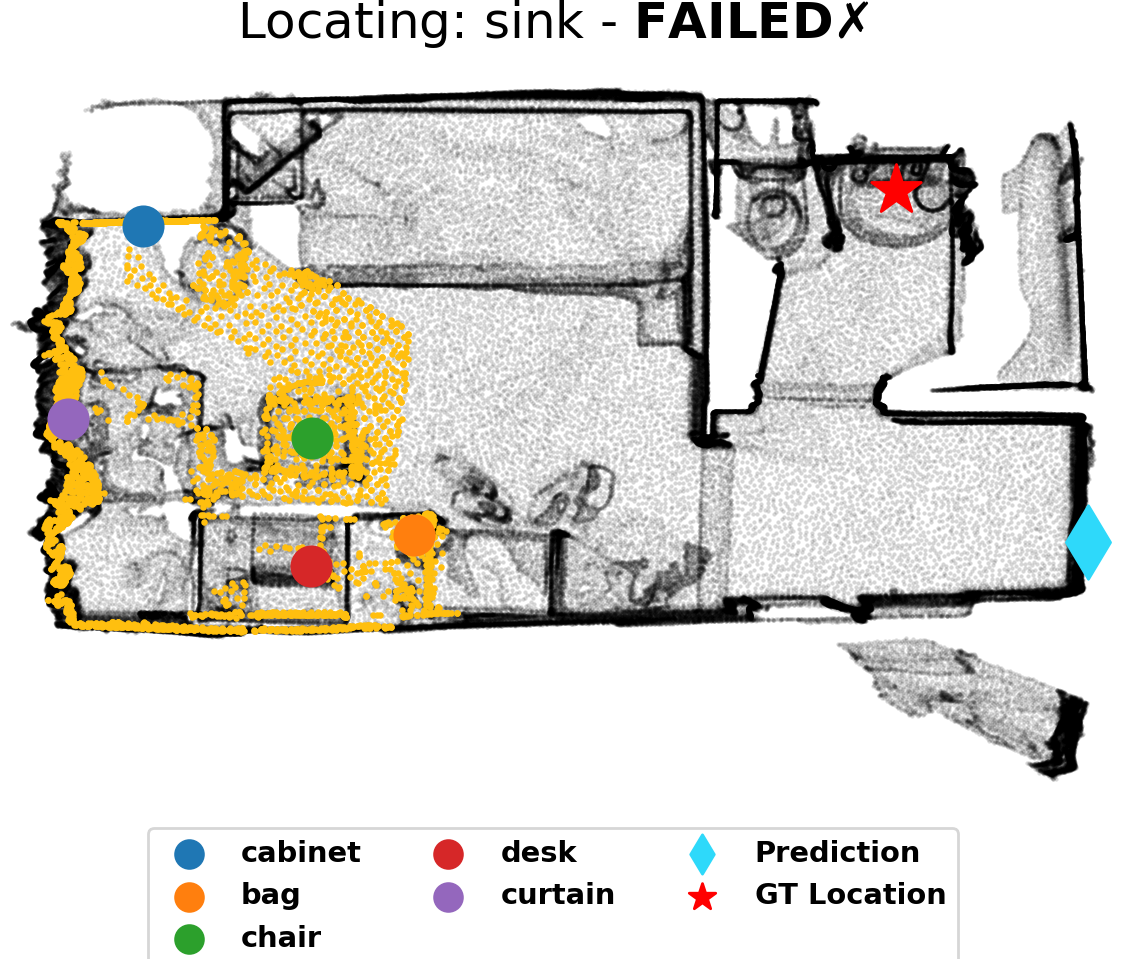}
    \includegraphics[height=0.3\textheight]{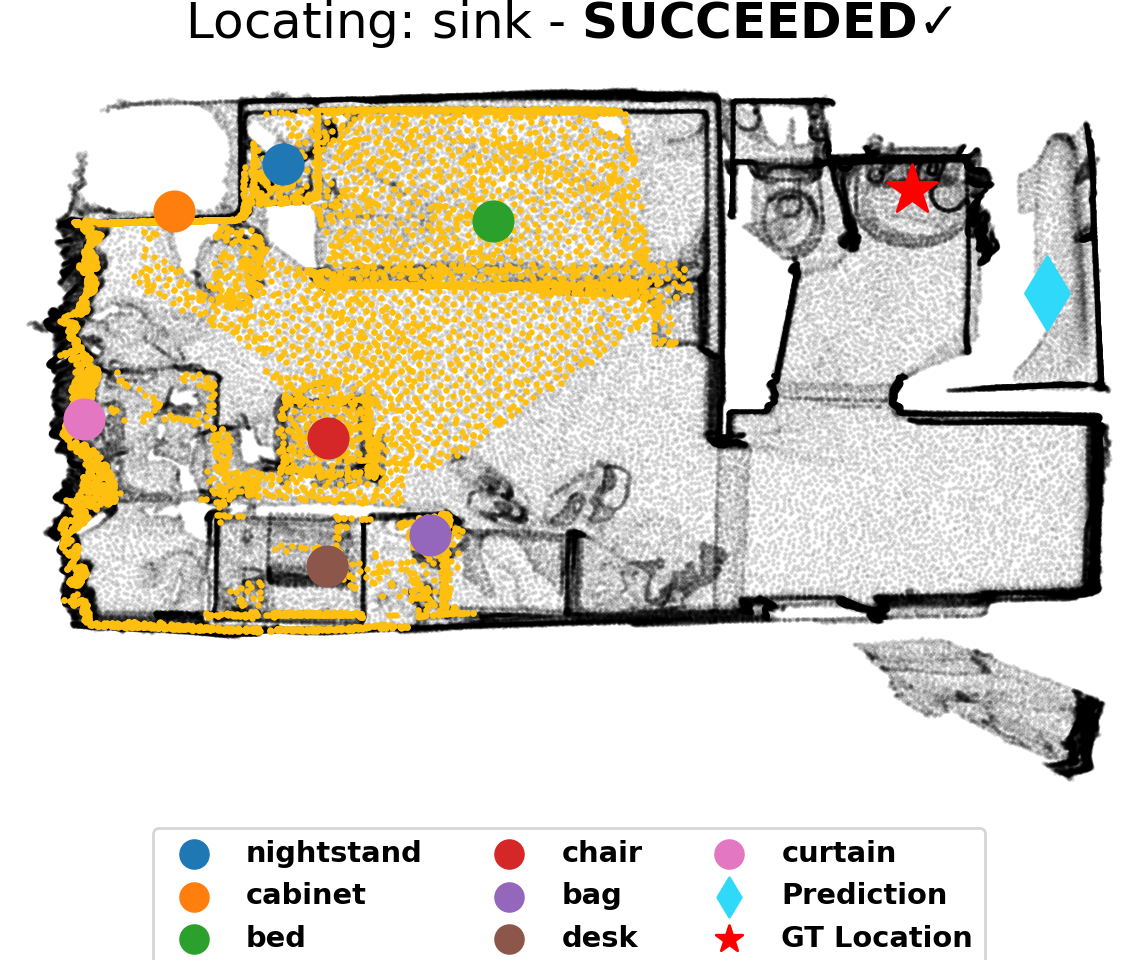}  \\
    \vspace{2em}
    \includegraphics[height=0.3\textheight]{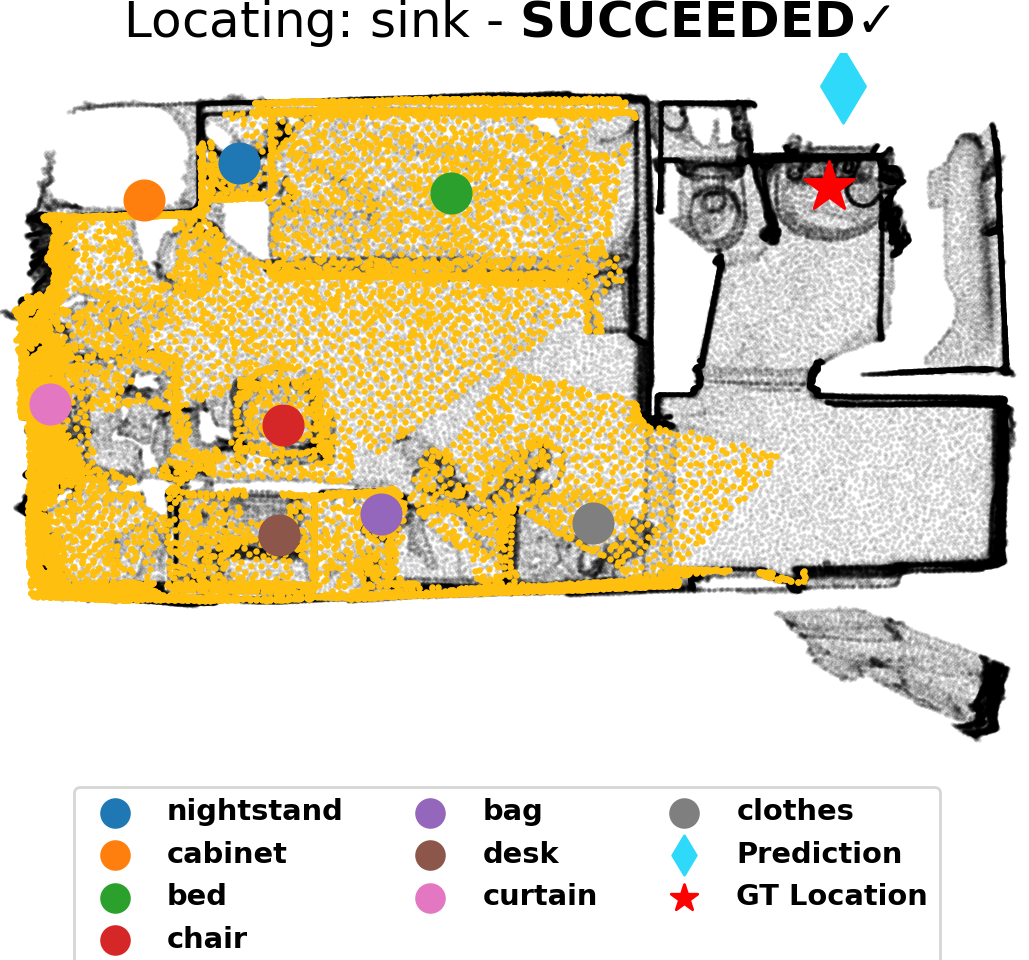} %
    \includegraphics[height=0.3\textheight]{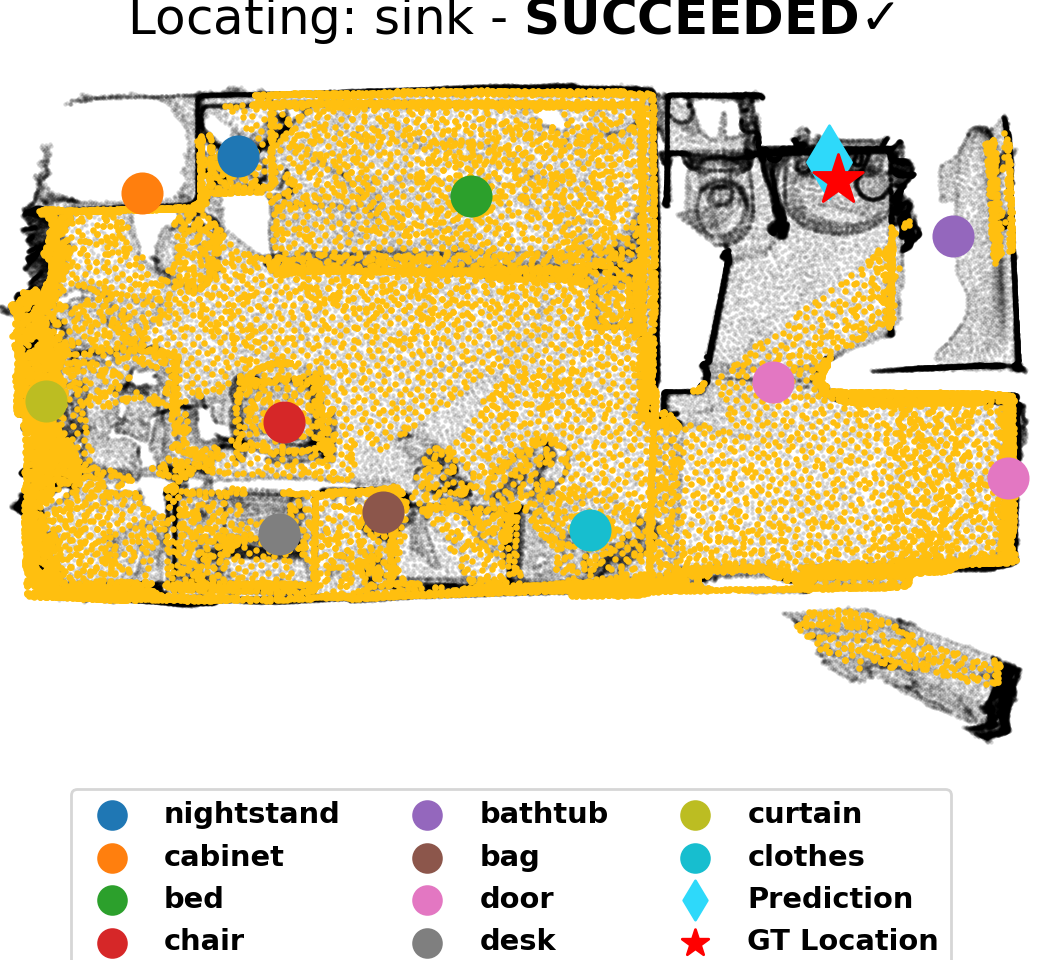}\\
    \caption{\textbf{Top} Complete scene of a small apartment. \textbf{Middle and Bottom} Localisation of the sink at different completeness levels. The localisation accuracy increases as the scene becomes more and more complete.}
    \label{fig:suppl:qualitative_sink}
\end{figure*}

\begin{figure*}[p]
    \centering
    \includegraphics[height=0.2\textheight]{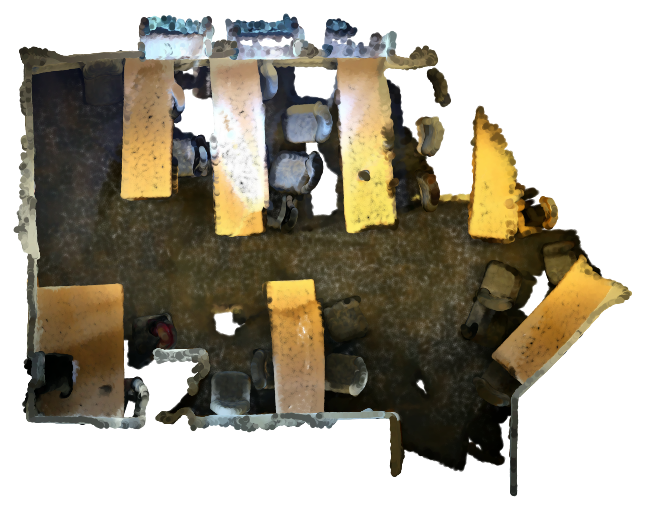} \\
    
    \includegraphics[height=0.3\textheight]{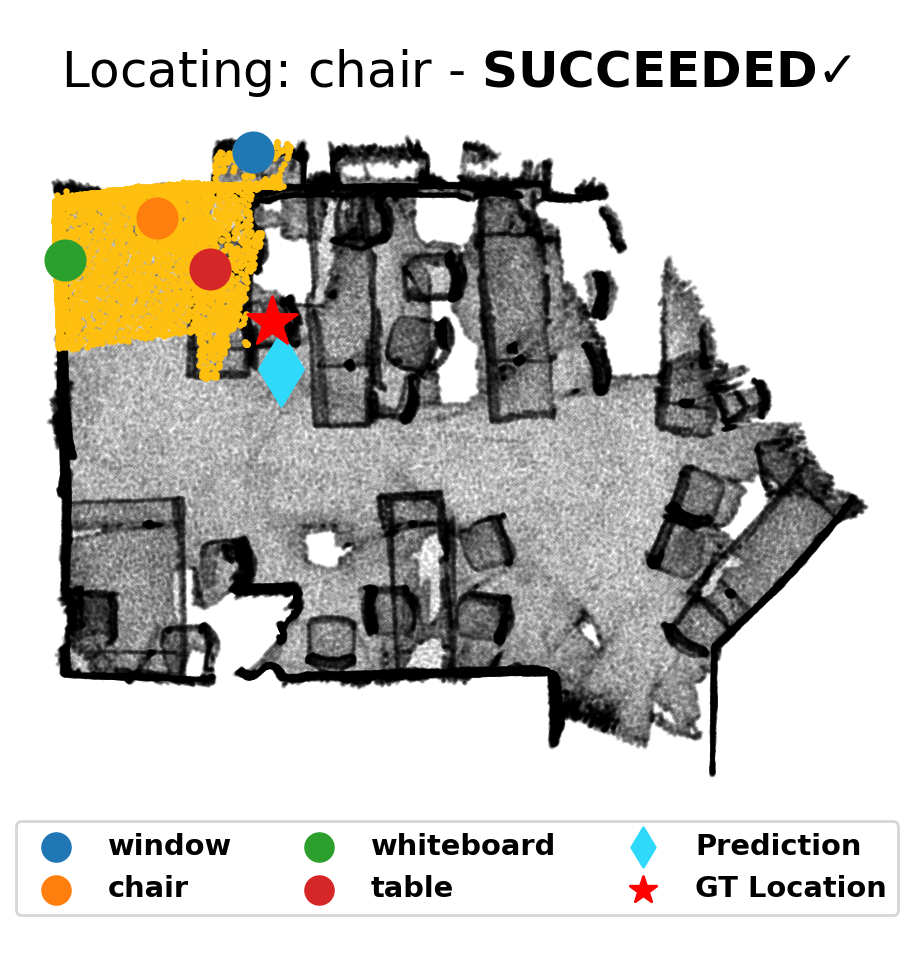} \hspace{6em}%
    \includegraphics[height=0.3\textheight]{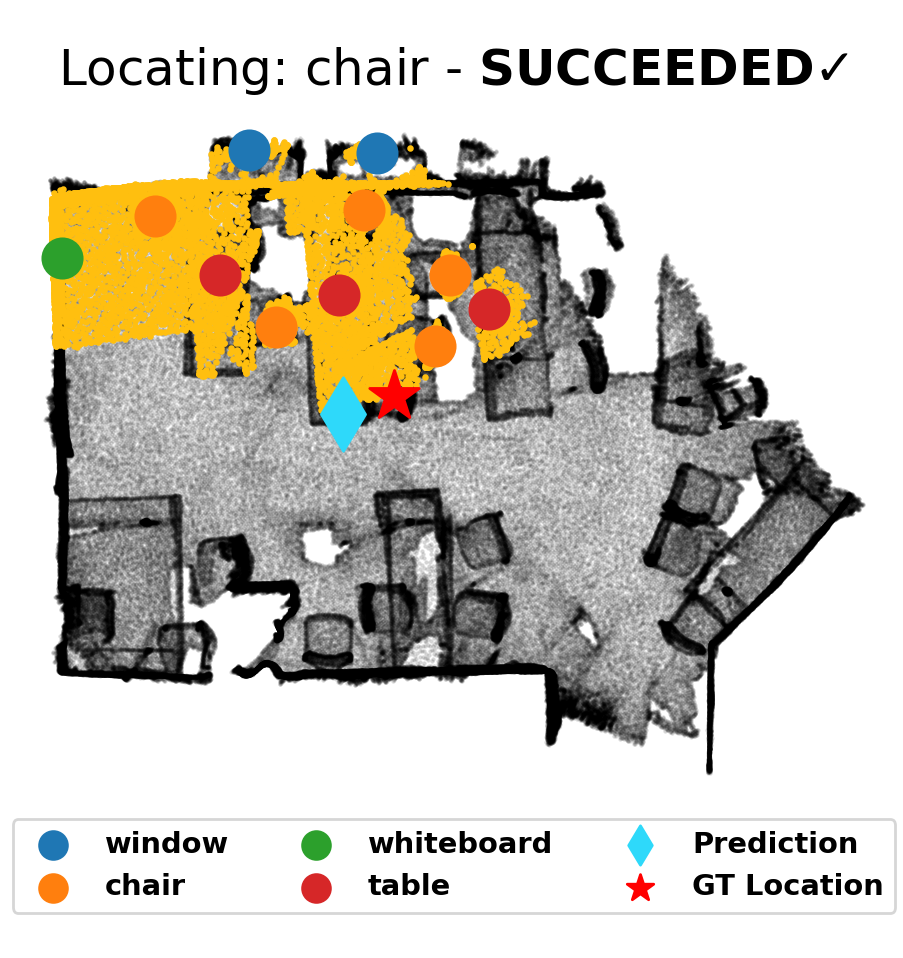}  \\
    \includegraphics[height=0.3\textheight]{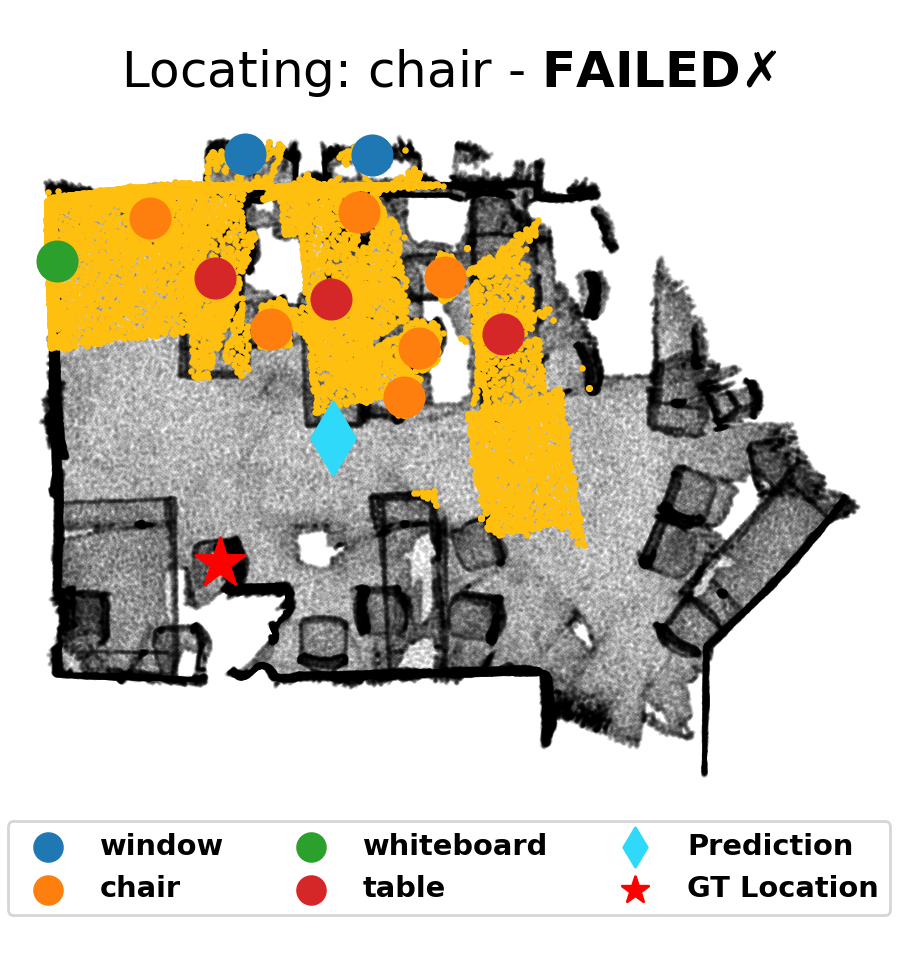} \hspace{6em}%
    \includegraphics[height=0.3\textheight]{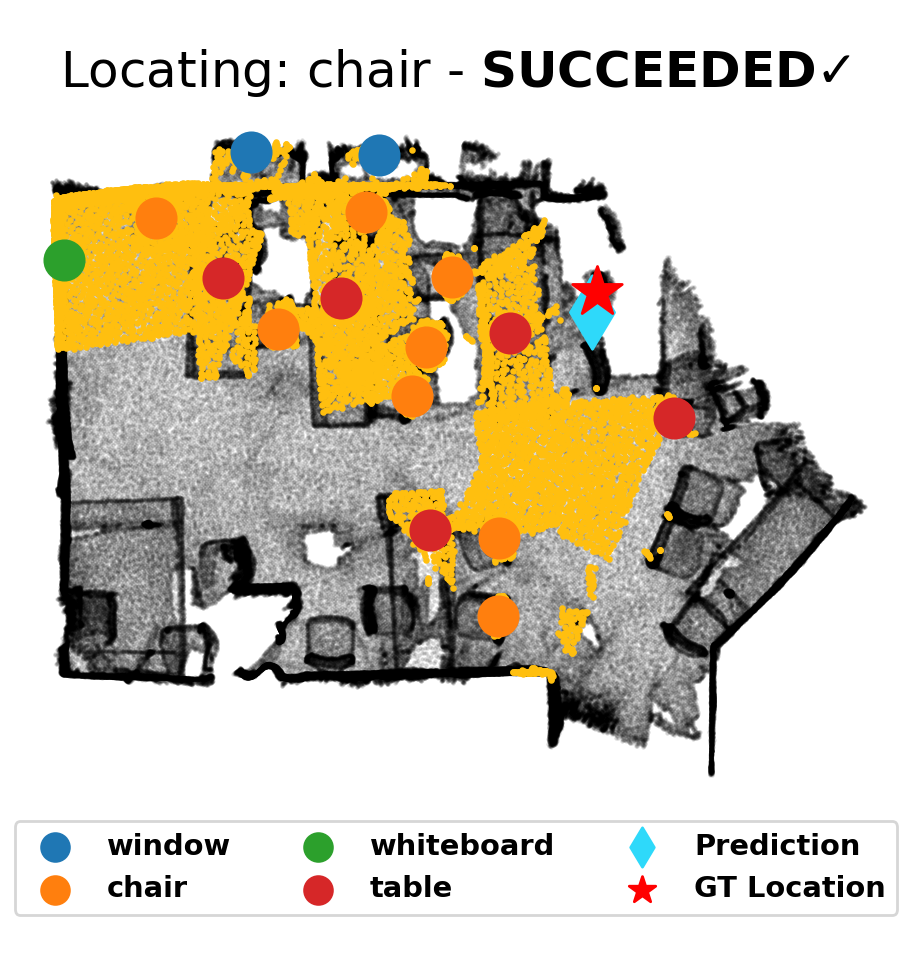}\\
    \caption{\textbf{Top} Complete scene of a classroom. \textbf{Middle and Bottom} Localisation of a chair at different completeness levels. Note that the red star indicates the ground-truth instance \textit{closest} to the prediction.  As the scene becomes more and more complete, the method is able to correctly adapt to changes in the SCG.}
    \label{fig:suppl:qualitative_class}
\end{figure*}

\noindent\textbf{Localisation at different levels of scenes completeness.} Figures \ref{fig:suppl:qualitative_sink} and \ref{fig:suppl:qualitative_class} show examples of localisation at different scene completeness levels. 
In the first case (Fig. \ref{fig:suppl:qualitative_sink}) we show the localisation of a sink in an apartment, while in the second case (Fig. \ref{fig:suppl:qualitative_class}) we show the localisation of a chair in a classroom.
In general, as the scenes become more complete (left to right), the predicted position gets closer to the ground-truth position. The qualitative results coincide with the quantitative results presented in Fig.~{\color{red}5} of the main paper, i.e. the localisation error decreases and the Localisation Success Rate (LSR) increases with the scene completeness.

Interestingly, where there are multiple instances of the target object category, i.e. the case in Fig. \ref{fig:suppl:qualitative_class}, we can see how the \mname\, localises different instances of a chair based on the completeness of the scene. 
When the scene is mostly unobserved, i.e. the top-left case, our method places the chair behind a table in front of a whiteboard and manages to locate it correctly. In the top-right case with a more complete scene, \mname\,correctly localises another chair on the side of the table. This is because the chair that was located in top-left case is now part of the SCG, thus not a valid target. 
In the bottom-left case, the previously predicted chairs are now part of the SCG, therefore no longer a valid target for the localisation. The network predicted a new position at the head of the table, although plausible, is considered a failure as there isn't a chair in the vicinity. 
In the bottom-right case, the model correctly predicts the most plausible position for a chair is between the two tables. 

\section{Ethical Discussion}
\label{sec:suppl_ethical}
Our new dataset has been built on top of the ScanNet dataset. Since ScanNet does not contain any human subject, by proxy, neither does our dataset of partial reconstructions.

Moreover, we proposed a novel graph modelling that enrich spatial scene representation with commonsense knowledge. Such formulation has a broader impact on the research community and foster methods for perception tasks that require spatial representation learning. In this paper, we demonstrated its effectiveness in terms of inferring the position of objects in unknown scenes, which by itself introduces potentials to advance applications, such as localisation service or suggestive layout design.

The proposed graph formulation aims to understand how we as humans model the arrangement of objects in rooms, and thus to learn a layout ``profile''. We note that this profile has been learned on thousands of different scenarios and therefore is too broad and generic to be used to negatively target specific individuals, races, or groups.

\end{document}